\documentclass[10pt]{article}

\usepackage{amsmath}
\usepackage{amssymb}

\usepackage{cite}

\usepackage{hyperref}

\usepackage{lineno} 

\usepackage{microtype}
\DisableLigatures[f]{encoding = *, family = * }

\usepackage{graphicx} 
\usepackage[CaptionAfterwards]{fltpage} 
\usepackage{float}
\usepackage{lipsum}

\usepackage[labelfont=bf,labelsep=period,justification=raggedright]{caption}

\makeatletter
\renewcommand{\@biblabel}[1]{\quad#1.}
\makeatother

\date{}

\newcommand{\ib}[1]{\mathbf{#1}}
\newcommand{\SFpef}{\ref{fig:SFpef}}
\newcommand{\SFexp}{\ref{fig:SFexp}}
\newcommand{\SFaln}{\ref{fig:SFaln}}
\newcommand{\SFinc}{\ref{fig:SFinc}}
\newcommand{\SFfda}{\ref{fig:SFfda}}
\newcommand{\SFenc}{\ref{fig:SFenc}}

\begin{document}

\begin{flushleft}
{\Large
\textbf{Measuring and Understanding Sensory Representations within Deep Networks Using a Numerical Optimization Framework}
}
\\
Chuan-Yung Tsai$^{1,2}$, 
David D. Cox$^{1,2,3,\ast}$
\\
\bf{1} Department of Molecular and Cellular Biology, Harvard University, Cambridge, MA, USA
\\
\bf{2} Center for Brain Science, Harvard University, Cambridge, MA, USA
\\
\bf{3} Department of Computer Science, Harvard University, Cambridge, MA, USA
\\
$\ast$ E-mail: davidcox@fas.harvard.edu
\end{flushleft}

\section*{Abstract}
A central challenge in sensory neuroscience is describing how the activity of populations of neurons can represent useful features of the external environment. However, while neurophysiologists have long been able to record the responses of neurons in awake, behaving animals, it is another matter entirely to say what a given neuron does. A key problem is that in many sensory domains, the space of all possible stimuli that one might encounter is effectively infinite; in vision, for instance, natural scenes are combinatorially complex, and an organism will only encounter a tiny fraction of possible stimuli. As a result, even describing the response properties of sensory neurons is difficult, and investigations of neuronal functions are almost always critically limited by the number of stimuli that can be considered.

In this paper, we propose a closed-loop, optimization-based experimental framework for characterizing the response properties of sensory neurons, building on past efforts in closed-loop experimental methods, and leveraging recent advances in artificial neural networks to serve as as a proving ground for our techniques. Specifically, using deep convolutional neural networks, we asked whether modern black-box optimization techniques can be used to interrogate the ``tuning landscape'' of an artificial neuron in a deep, nonlinear system, without imposing significant constraints on the space of stimuli under consideration. We introduce a series of measures to quantify the tuning landscapes, and show how these relate to the performances of the networks in an object recognition task. To the extent that deep convolutional neural networks increasingly serve as \emph{de facto} working hypotheses for biological vision \cite{yamins2014performance, cadieu2014deep, khaligh2014deep}, we argue that developing a unified approach for studying both artificial and biological systems holds great potential to advance both fields together.


\section*{Introduction} 
\label{sec:intro}
The mammalian ventral visual pathway consists of a hierarchical cascade of visual areas that progressively reformat visual information from pixel-like retinotopic representations into formats suitable for high-level tasks \cite{dicarlo2007untangling, dicarlo2012does}, such as recognizing visual objects. 
Recently, computer vision algorithms inspired by the hierarchical structure of the visual system---deep convolutional neural networks \cite{fukushima1980neocognitron, lecun1998gradient, riesenhuber1999hierarchical}, or so-called ``deep learning'' approaches \cite{krizhevsky2012imagenet}---have become the focus of tremendous attention in the computer vision and machine learning communities, due to their success in a variety of practical domains. 
In particular, these networks can perform on par with human subjects on certain controlled visual tasks \cite{serre2007feedforward, cirecsan2012multi, russakovsky2014imagenet, taigman2013deepface, sun2014deep, viglarge} and they produce internal representations that are similar to that of mammalian visual systems under certain conditions \cite{yamins2014performance, cadieu2014deep, khaligh2014deep}.
However, such networks are still far from achieving human-level capabilities for unconstrained visual tasks \cite{ghodrati2014feedforward}, and even for relatively constrained object categorization tasks, surprising failures can be induced through subtle stimulus manipulation \cite{szegedy2013intriguing}.
More fundamentally, we lack a comprehensive functional understanding of how these system work at a theoretical level. Here, we take the success of artificial ``deep learning'' architectures as an opportunity to develop tools for studying and gaining insight about the function of deep networks of nonlinear units---tools that can in principle be used to study both artificial and biological neuronal systems.

Explaining the sensory representations of neurons, one of the most important aspects of studying the underpinnings of sensory processing circuitry, has been a widely researched topic in both theoretical and experimental neuroscience, where artificial and biological neural networks are mainly used and often cross-studied for developing better explanations of sensory processing mechanisms.
Brief reviews of methods used in both fields are as follows.
(1) In theoretical neuroscience studies, methods can be categorized by the artificial models being used \cite{wu2006complete}: parametric or nonparametric (which are usually analytically or only numerically analyzable).
Berkes et al.~\cite{berkes2006analysis} studied quadratic networks, $f\left(\ib{x}\right) = \frac{1}{2}\ib{x}^{T}\ib{Qx}+\ib{L}^{T}\ib{x}+c$, in which the optimal stimulus uniquely exists and can be efficiently computed, and local invariance and selectivity directions can be analytically derived through eigendecomposition of the symmetric quadratic term $\ib{Q}$ (i.e.~the Hessian), which are eigenvectors corresponding to the least and most negative eigenvalues.
Saxe et al.~\cite{saxe2011random} studied single-level convolutional networks $f(\ib{x}) = \left\| \ib{K} \otimes \ib{x} \right\|_{F}$, which can be viewed as the building block of deep convolutional networks, and showed the optimal stimulus can be analytically approximated using a Gabor-like filter with its frequency component collocated with the peak of the Fourier spectrum of convolution kernel $\ib{K}$, whether $\ib{K}$ itself is structural or random.
Zeiler et al.~\cite{zeiler2014visualizing} and Simonyan et al.~\cite{simonyan2013deep} studied multi-layer convolutional networks, numerically derived and approximated the optimal stimuli of neurons of various depths, and qualitatively showed in deeper layers, neurons are tuned to gradually more complex visual patterns.
Zeiler et al.~\cite{zeiler2014visualizing} also used parametric deformations (translation, rotation, and scaling) to test the invariance of neurons.
Although methods in \cite{zeiler2014visualizing, simonyan2013deep} didn't require the networks being fully analytical, being able to perform backpropagation-based gradient estimation in the networks is however needed.
Le et al.~\cite{ngiam2010tiled} extended the method in \cite{berkes2006analysis} onto multi-layer convolutional networks through numerically estimating the optimal stimulus and the Hessian.
Le et al.~\cite{le2012building} also visualized the optimal stimulus of multi-layer convolutional networks, but unlike \cite{zeiler2014visualizing, simonyan2013deep}, treated the networks as non-analytical black boxes.
Erhan et al.~\cite{erhan2010understanding} studied multi-layer networks and proposed to characterize the invariance in a larger extent, through numerically searching for non-local (i.e.~farther to the optimal stimuli) solutions, instead of only locally estimating it through Hessian decomposition.
(2) In experimental neuroscience studies, methods can be categorized by the stimulus being used to characterize the biological models \cite{wu2006complete}: parametric or nonparametric, as well.
An $N$ dimensional parametric stimulus $\ib{x}$ can be either functionally generated through a $P$ dimensional latent parameter $\ib{p}$, i.e.~$\ib{x}\left(\ib{p}\right) \in \mathbb{R}^N$ and $\ib{p} \in \mathbb{R}^P$ where $P < N$, or sampled from a stimulus dictionary of limited size that mostly only spans part of the $N$ dimensional space \cite{field1987relations}; e.g., moving bars, sine gratings, natural images, etc.~fall into this category.
It is an efficient and commonly adopted stimulus for studying sensory representations of neurons in various stages along the sensory pathways, and can be coupled with characterization procedures ranging from closed-loop methods like genetic algorithm \cite{bleeck2003using, yamane2008neural}, to open-loop methods like spike-triggered average (i.e.~reverse correlation) \cite{ringach2004reverse, hansen2004parametric, dotsch2012reverse} and Bayesian methods \cite{naselaris2009bayesian, nishimoto2011reconstructing}.
Contrarily, an $N$ dimensional nonparametric stimulus $\ib{x} \in \mathbb{R}^N$ can have all its variables set independently and thus spans the entire $N$ dimensional space.
White noise is the most commonly used modality, which can be coupled with characterization procedures ranging from closed-loop methods like hill climbing \cite{harth1974alopex}, to open-loop methods like spike-triggered covariance \cite{touryan2002isolation, rust2004spike} and spike-triggered average as well \cite{ringach2004reverse}.
Review of methods focusing on closed-loop characterization also can be found in \cite{dimattina2013adaptive}. 


In this work, we propose a unified framework based on modern numerical optimization techniques to help us uncover the sensory representations encoded by neurons deep inside neural networks, where the most flexible and generalizable settings---nonparametric network model with nonparametric stimulus---are supported to maximize the applicability of this framework, with the inherent difficulties and inefficiency resolved by carefully constructing and constraining the numerical framework.
Characterization of the ``tuning landscapes'' of neurons to the high-dimensional nonparametric stimulus includes both first- and quasi-second-order ``landscape features'', i.e.~optimal stimulus and its invariance and selectivity directions (arguably the most significant structural features of the surrounding landscape, and thus, the inefficiency of modeling the entire Hessian can be avoided).
Through incorporating multiple randomized searches, the invariance and selectivity subspaces can be efficiently estimated as well.
A series of representation measures for analyzing the numerical search results are also designed to perform dimensionality-insensitive characterization of deep networks.
Using the proposed framework for sensory representation characterization, we directly {addressed} two important questions: (1) Why are deep networks better than shallow networks? (2) Among experimented networks of the same depth, why are certain networks better than the others? 

\section*{Methods} 
\label{sec:methods}
\begin{figure}
\centering \includegraphics[width=0.5\textwidth]{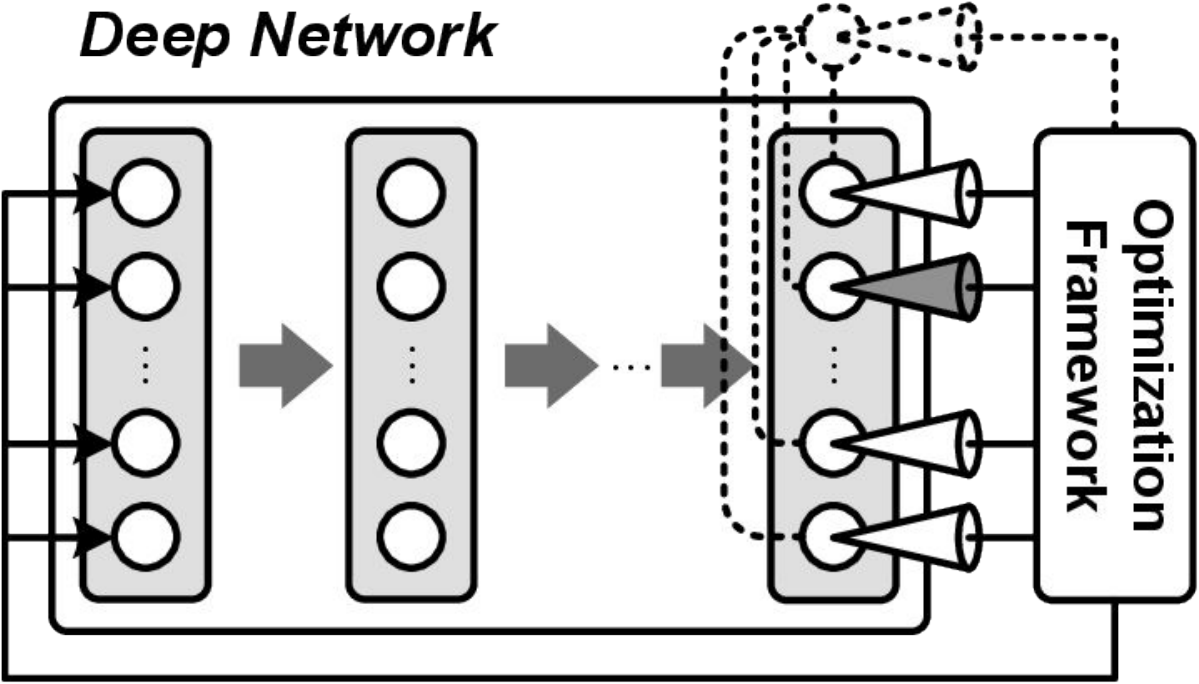} 
\caption{ 
{\bf Schematic of numerical optimization framework.} Deep network $f$: hierarchical cascade of layers (shaded tall rectangles) of artificial neurons (circles) transforming the stimulus $\ib{x}$ (bottom layer, leftmost) into high-level representations $\ib{r}$ (top layer, rightmost) suitable for linear classification. Optimization framework: closed-loop algorithm that repeatedly measures the response(s) of one or multiple neurons, calculates the fitness, and adaptively generates stimuli to characterize the target neuron(s) via optimizing the fitness functions defined for different purposes (see Eq.~(\ref{eq:O1}--\ref{eq:S2})); when characterizing multiple neurons for population representation, response of an imaginary neuron (dashed circle) tuned to the given optimal or reference stimulus is optimized (see Eq.~(\ref{eq:O2}, \ref{eq:I2}, \ref{eq:S2})) equivalently.}
\label{fig:concept}
\end{figure}

Figure \ref{fig:concept} illustrates and defines the setup of the closed-loop numerical optimization framework for studying sensory representations encoded within the deep network of interest. The proposed framework consists of two main methods: (1) optimal stimulus search and (2) invariance/selectivity path search, or the first-order (i.e.~linear) and quasi-second-order (i.e.~low-rank quadratic, where only eigenvectors corresponding to the largest and smallest eigenvalues are considered) characterization methods. Representation $\ib{r}$, in this work, is equivalent to the response(s) of artificial neuron(s), and is denoted as unit representation (or scalar representation, i.e.~$\ib{r} \in \mathbb{R}$) when referring to single neurons and population representation (or vector representation, i.e.~$\ib{r} \in \mathbb{R}^R$) when referring to groups of neurons (with group size $R$). Arguably, for studying both artificial and biological neural networks, especially when evaluating task-related performances, the viability of characterizing population representations is usually more important than of characterizing unit representations, considering the differences in representational powers. This framework supports characterizing both unit (using Eq.~(\ref{eq:O1}, \ref{eq:I1}, \ref{eq:S1}), though Eq.(\ref{eq:O2}, \ref{eq:I2}, \ref{eq:S2}) can be used as well) and population (using Eq.(\ref{eq:O2}, \ref{eq:I2}, \ref{eq:S2})) representations in a unified way.

As aforementioned, this framework treats networks as non-analytical black boxes---it makes no assumption about networks' structures or properties (e.g.~we do not assume that analytical gradients are available, even though they do exist in most artificial networks). This strategic decision {was} made in order to allow the proposed methods to transfer over to experiments with biological neurons, where only the outputs of neurons are available. While no specific knowledge of the network itself is assumed, we do restrict the space of stimuli considered, based on prior knowledge of biological and artificial neuronal response properties. In particular, all stimuli considered in this work follow a constant energy constraint $\left\| \ib{x} \right\| = E$, complying with the general observation that neurons are often modulated by stimulus contrasts \cite{albrecht1982striate, cheng1994comparison}, but this modulation is less interesting when considering pattern selectivity of neurons. Limiting the stimulus search space in this manner effectively reduces the range of possible stimuli to consider, and avoids degenerate solutions that simply maximize stimulus contrasts. For simplicity of mathematical formulations, the setting of $E=1$ is used for the rest of the paper, while in the experiments $E$ {was} set to the average energy of task-related stimuli.

In all experiments, numerical optimizations {were} executed multiple times starting from different random initial stimuli for two main reasons: (1) increasing robustness/quality of the numerical solutions of, e.g., optimal stimulus and invariance/selectivity path searches (2 runs executed), and (2) providing statistical samplings of the solution spaces of, e.g., encoding specificity and invariance/selectivity subspace searches (10 and 20 runs executed). Worth to note, since the output responses (i.e.~tuning landscapes) of artificial neural networks are in general non-convex, there is no guarantee that optimality can be achieved through any optimization method. However, through carefully constructing the numerical solver and performing multiple runs of optimization, we argue that informative local optima can still be discovered and used to characterize the unit/population representations.


\subsection*{First-Order Characterization: Optimal Stimulus Search and Analysis}

\begin{figure}
\centering \includegraphics[width=0.5\textwidth]{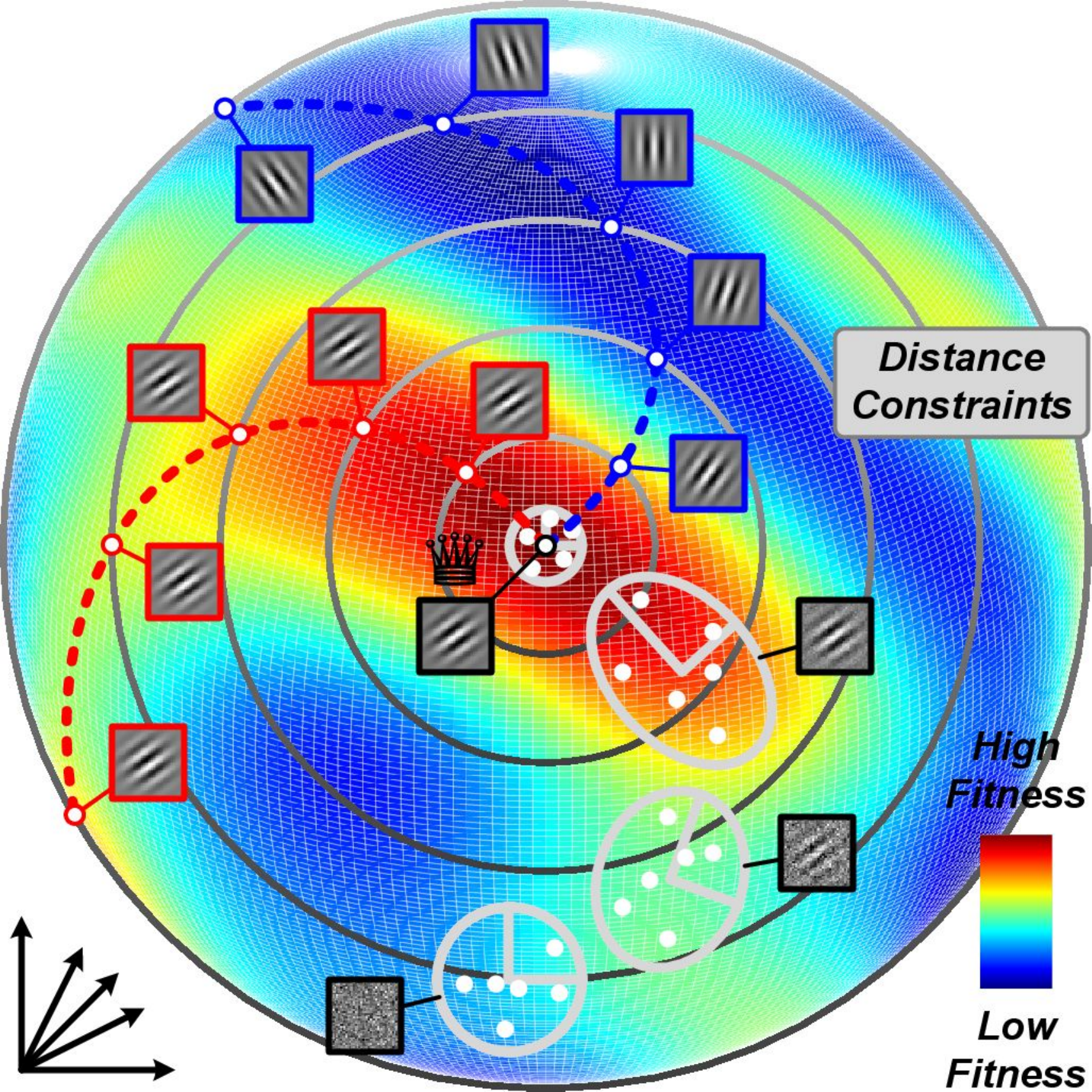} 
\caption{
{\bf Visualization of numerical optimization procedures.} Spherical constraint: solution space of the stimulus with energy constraint $\left\| \ib{x} \right\| = 1$, an $N$ dimensional sphere centered on the origin. Optimal stimulus search trajectory: best solutions (patterns with black frame boarders) from sequential search iterations with adaptive landscape modeling (gray eclipses with varying diameters). Distance constraints (gray circles): solution spaces of stimuli with $0.1\pi$ to $0.5\pi$ distances from the optimal stimulus. Invariance path (dashed red curve): solutions of the invariance path searches. Selectivity path (dashed blue curve): solutions of the selectivity path searches. Target neuron in this visualization is optimally tuned to a $45^{\circ}$ Gabor filter, invariant to phase changes, and selective to orientation changes.}
\label{fig:methods}
\end{figure}

In this work, optimal stimulus is numerically derived through the iterative optimization as \begin{equation} \label{eq:O1} \hat{\ib{x}} = \underset{\ib{x}} {\arg\max} f\left(\ib{x}\right) = \underset{\ib{x}_{g} \in \Omega \left( \ib{x}_{g-1} , \mathcal{M}_{g-1} \right)}{\arg\max} f\left(\ib{x}_{g}\right),\end{equation} where $\mathcal{M}_{g}=U\left(\left\lbrace f\left(\ib{x}_{g}\right) \right\rbrace , \mathcal{M}_{g-1}\right)$ and $1 \le g \le G$, subject to $\left\| \ib{x} \right\| = 1$. The algorithm starts from random initial point $\ib{x}_{0}$ on an $N$ dimensional sphere, as the noisy pattern shown in the bottom of Fig.~\ref{fig:methods}, and samples a set of $\lambda$ neighboring points (through function $\Omega$) as candidates to be evaluated, from an initial model $\mathcal{M}_{0}$ with null distribution (in this work, multivariate Gaussian distribution starting with covariance matrix $\boldsymbol{\Sigma} = \ib{I}$). Through measuring the fitnesses of sampled points $\left\lbrace f\left(\ib{x}_{g}\right) \right\rbrace$, the algorithm updates its model $\mathcal{M}_{g}$ (i.e.~mean and covariance of the distribution, through function $U$) for generating samples in the following iterations, where the same operations repeat. Usually the maximally allowed iteration number $G$ is set to prevent the algorithm from running unreasonably long; nevertheless, early termination is also common in practice when the neighborhood size shrinks below threshold, indicating that optimality is reached as shown in the center of Fig.~\ref{fig:methods}. Though the fitness landscape can be highly nonlinear, such sequential optimization procedure is usually capable of gradually accumulating knowledge of the landscape and adjusting its search directions to climb onto higher fitness areas, and the smoother the landscape is, the faster the algorithm converges. As exemplified in Fig.~\ref{fig:methods}, the ``most salient pattern'', or direction leading to high fitness areas usually can be rapidly extracted (see Fig.~\ref{fig:ind_res} for more examples). In most scenarios, fixing the maximally allowed number of fitness evaluations $\lambda G$ to, e.g., $100N$ (as adopted for the optimal stimulus search in this work) can lead to reasonably good convergence speed vs.~quality trade-off. The aforementioned sequential optimization concept is in fact commonly adopted in modern stochastic optimization \cite{spall2005introduction}, and the CMA-ES (Covariance Matrix Adaptation Evolution Strategy) algorithm \cite{hansen2001completely} is chosen as the back-end solver of this numerical framework, for its decent convergence speed and capability of handling rugged landscapes. Readers can refer to \cite{hansen2001completely} for detailed definitions of model $\mathcal{M}$ and functions $\Omega$ and $U$. The energy constraint of stimulus is simply handled through a spherical projection before stimulus evaluation, i.e.~$f\left(p_s\left(\ib{x}\right)\right)$ where $p_{s}\left(\ib{x}\right) = \ib{x} / {\left\|\ib{x}\right\|}$, while the solver works unconstrainedly.

For analyzing the results of optimal stimulus searches, the following measures are adopted: (1) \emph{Spectral complexity}, which is estimated through the $L^{1}$ norm of the (2 dimensional) Fourier power spectrum of optimal stimulus, i.e.~$\left\| \mathcal{F}(\hat{\ib{x}}) \right\|_{1}$, where higher value suggests higher non-sparsity (i.e.~more spectral components required to represent the signal). (2) \emph{Explanation power}, which is estimated as the mean of the ``linear explainability'' of task-related stimuli---rectified inner-product distances between a set of $n$ task-related stimuli $\left\lbrace \ib{x}^{t} \right\rbrace$ and the optimal stimulus itself, i.e.~$\frac{1}{n}\sum_{i=1}^{n}\max\left(\langle \ib{x}^{t}_{i} , \hat{\ib{x}} \rangle , 0\right)$---and measures how well the optimal stimulus can linearly ``approximate'' task-related stimuli (or intuitively how much neuronal response might be elicited). (3) \emph{Encoding specificity}, which utilizes the optimal stimulus search as an inverse function $f^{-1}$ of a representation $\ib{r}$ (i.e.~searching for the optimal stimulus of an imaginary neuron, as shown in Fig.~\ref{fig:concept}, tuned to the representation of a reference stimulus) and measures the average of structural similarities (SSIM) \cite{wang2004image} between task-related reference stimulus $\ib{x}^{*}$ and $n$ reconstructed stimuli (with randomized initializations), i.e.~$\frac{1}{n}\sum_{i=1}^{n} \mathrm{SSIM}\left(\ib{x}^{*} , \hat{\ib{x}}^{*}_{i} \right)$ where \begin{equation} \label{eq:O2} \hat{\ib{x}}^{*} = \underset{\ib{x}}{\arg\max} \left( e^{-\left\|f\left(\ib{x}\right)-\ib{r}\right\|} \right) \end{equation} and $\ib{r} = f\left(\ib{x}^{*}\right)$, indicating how specific (or non-confounding) a representation $\ib{r}$ is encoded.


\subsection*{Quasi-Second-Order Characterization: Invariance and Selectivity Path Search and Analysis}

\begin{figure}
\centering \includegraphics[width=0.5\textwidth]{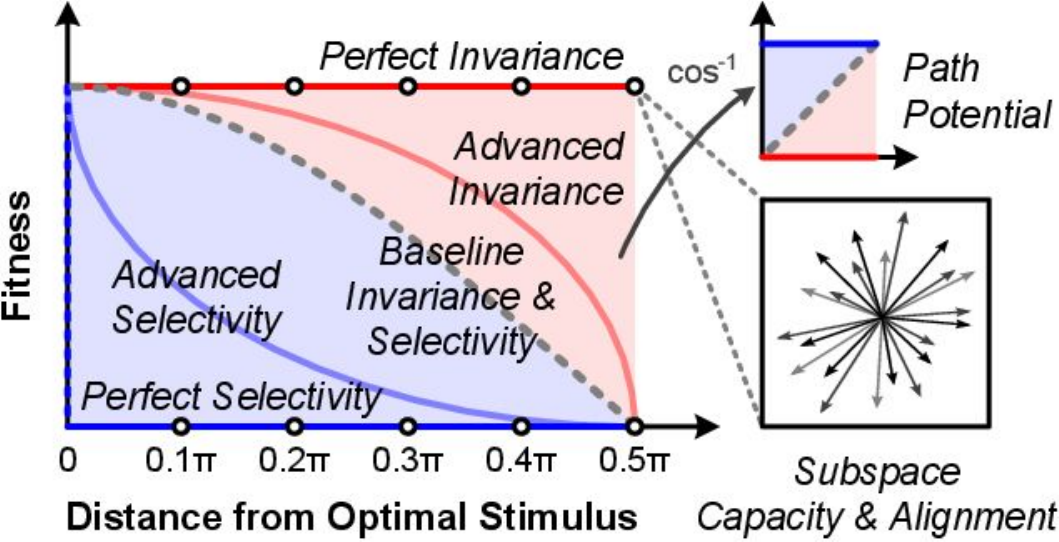} 
\caption{
{\bf Fitness-distance diagram.} Invariance (red) and selectivity (blue) curves: fitnesses of invariance and selectivity path search results plotted against distances from the optimal stimulus. Baseline (dashed gray) curve: graph of cosine function; invariance and selectivity curves of single inner-product neuron. Subsequent analyses, including path potential, subspace capacity and subspace alignment, can be performed. Invariance and selectivity path potentials can be defined on either the arccosine normalized diagram (shaded red and blue areas), or on the original diagram (line integrals).}
\label{fig:fd_diag}
\end{figure}

With respect to the optimal stimulus $\hat{\ib{x}}$, the searches of invariance and selectivity paths, which consist of sets of invariant stimuli ($\left\lbrace \ib{x}^{+}_{\delta} \right\rbrace$, i.e.~optimally excitatory) and selective stimuli ($\left\lbrace \ib{x}^{-}_{\delta} \right\rbrace$, i.e.~optimally inhibitory) respectively, are formulated as
\begin{align}
\ib{x}^{+}_{\delta} &= \underset{\ib{x}_{\delta}}{\arg\max} f\left(\ib{x}_{\delta}\right) \label{eq:I1} \\
&= \underset{\ib{x}_{\delta}}{\arg\max} \left( e^{-\left\|f\left(\ib{x}_{\delta}\right)-f\left(\hat{\ib{x}}\right)\right\|} \right); \label{eq:I2} \\
\ib{x}^{-}_{\delta} &= \underset{\ib{x}_{\delta}}{\arg\min} f\left(\ib{x}_{\delta}\right) \label{eq:S1} \\
&= \underset{\ib{x}_{\delta}}{\arg\min} \left( e^{-\left\|f\left(\ib{x}_{\delta}\right)-f\left(\hat{\ib{x}}\right)\right\|} \right); \label{eq:S2}
\end{align}
where $0 < \delta \le \frac{\pi}{2}$, subject to $\left\| \ib{x}_{\delta} \right\| = 1$ and $\langle \ib{x}_{\delta} , \hat{\ib{x}} \rangle = \cos\left(\delta\right)$. As visualized in Fig.~\ref{fig:methods}, the invariance path characterizes the $N$ dimensional ``curve'' that leads toward (or ideally maintains at) fitness as high as possible while moving away from the optimal stimulus, and the selectivity path characterizes the ``curve'' that leads toward fitness as low as possible while moving away. The search process is implemented as multiple runs of maximization/minimization on discretized $\delta \in \left\lbrace 0.1\pi, 0.2\pi, 0.3\pi, 0.4\pi, 0.5\pi\right\rbrace$, as the circular distance constraints shown in Fig.~\ref{fig:methods}, where each run is initialized with the result from previous run (and the $0.1\pi$ run directly with optimal stimulus $\hat{\ib{x}}$) to increase the path continuity and searching speed. Each $\delta$ of both paths has a budget of $20N$ fitness evaluations which in this work appears to be practically sufficient. The maximization and minimization simply use the same back-end solver, and the linear constraint $\langle \ib{x}_{\delta} , \hat{\ib{x}} \rangle = \cos\left(\delta\right)$ can be easily handled through an extra conic projection before stimulus evaluation, i.e.~$f\left(p_c\left(p_s\left(\ib{x}\right)\right)\right)$ where $p_c\left(\ib{x}\right) = \cos\left(\delta\right)\hat{\ib{x}} + \sin\left(\delta\right)\dot{\ib{x}} \mathbin{/} \left\|\dot{\ib{x}}\right\|$ and $\dot{\ib{x}} = \ib{x} - \langle\hat{\ib{x}},\ib{x}\rangle \hat{\ib{x}}$. The way the simple linear constraint is constructed to enforce the exploration of a larger extent of the fitness landscape is one of the main differences compared to \cite{erhan2010understanding}. In this work, the distance constraint $\delta$ only goes up to $\frac{\pi}{2}$, since on the $N$ dimensional sphere, stimuli fall in $\frac{\pi}{2} < \delta \le \pi$ are simply ``negatives'' of those in $0 < \delta \le \frac{\pi}{2}$, which are of less uniqueness and interest; nevertheless, going up to the full range $0 < \delta \le \pi$ is numerically supported.

For analyzing the results of invariance/selectivity path searches, the following measures are adopted: (1) \emph{Path potential}: while the search results of invariance/selectivity paths can be visualized via the \emph{fitness-distance diagram} \cite{jones1995fitness} as shown in Fig.~\ref{fig:fd_diag}, where perfect invariance and selectivity are flat lines at highest and lowest fitnesses respectively, the baselines of invariance and selectivity also can be intuitively (and analytically) defined as paths of the simplest form of neural networks---an inner-product neuron, $f(\ib{x}) = \ib{w}^{T}\ib{x}$---which precisely overlap and follow the monotonic cosine falloff from its optimal stimulus $\hat{\ib{x}}=\ib{w}$, as $f(\ib{x}_{\delta}) = \cos(\ib{x}_{\delta})$ by definition; the invariance (and selectivity, similarly) path potential of a unit representation can thus be defined as $\int_{0}^{\frac{\pi}{2}}{\left| \cos^{-1}\left(f\left( \ib{x}^{+}_{\delta} \right)\right) - \delta \right|}\mathrm{d}\delta \mathbin{/} {\frac{\pi}{2}}$, the area sandwiched between the invariance (and selectivity) and baseline curves (in $\cos^{-1}$ domain such that both potential values fall in the range $\left[0,1\right]$), measuring how invariant (and selective) a target neuron is compared to the baseline (i.e.~zero invariance/selectivity potential); for population representation where no baseline can be easily identified, the invariance (and selectivity) path potential is alternatively defined as $\int_{0}^{\frac{\pi}{2}}{\exp\left(-\left\|f\left(\ib{x}^{+}_{\delta}\right)-f\left(\hat{\ib{x}}\right)\right\|\right)}\mathrm{d}\delta \mathbin{/} {\frac{\pi}{2}}$. (2) \emph{Subspace capacity}: compared to path potential, which is designed to characterize the best invariance/selectivity path even when only one of such exists, subspace capacity estimates the ``dimensionality'' (i.e.~how diverse different paths can be) of the linear subspace formed by multiple path search results via the nuclear norm of concatenation of $n$ results $\left\|\left[\ib{x}_{\delta,1},\dots,\ib{x}_{\delta,n}\right]\right\|_{*}$ (in this work $n=20$ runs at $\delta=0.1\pi$). (3) \emph{Subspace alignment}, which measures the alignment between subspaces formed by task-related stimuli and invariance/selectivity paths (i.e.~how likely the invariance/selectivity can benefit, e.g., stimulus recognition) via estimating the sparsity of $n$ path search results projected onto the principal component vectors $\ib{V}$ of task-related stimuli $\left\lbrace \ib{x}^{t} \right\rbrace$, i.e.~$\frac{1}{n}\sum_{i=1}^{n} \left\| \ib{Vx}_{\delta,i} \right\|_{1}$ ($n=20$ runs at $\delta=0.1\pi$ as well). 

\section*{Results} 
\label{sec:results}

The methods and measures proposed in this work {were} tested using the Spatio-Temporal Hierarchical Object Representation (STHOR) network \cite{pinto2009high, sthor}, a simplified variety of deep convolutional neural networks \cite{fukushima1980neocognitron, lecun1998gradient, riesenhuber1999hierarchical, krizhevsky2012imagenet}. Although, like other deep networks, it also consists of the standard cascade of convolution, nonlinear activation, pooling, and normalization layers (which together define a single ``level'' in this work), its convolution kernel weights are however randomly assigned rather than trained. While it doesn't match the performances of deep networks trained on massive quantities of data \cite{krizhevsky2012imagenet}, it can in fact be surprisingly competitive for a variety of tasks, especially in the regime where large quantities of training samples are not available \cite{pinto2009high, cox2011beyond, viglarge}. Most importantly, since it doesn't require training (except for the linear SVM classifier on top of the network), large numbers of networks with diverse structural parameters can be rapidly generated and compared, which is particularly useful in this study where relative, instead of absolute task performances, are of the main interest. In this work, 100 shallow networks (i.e.~one-level, or L1, models) and 100 deep networks (i.e.~two-level, or L2, models) {were} randomly generated (all with 32 top-layer neurons) and tested to see how their representations vary with networks' depths and affect networks' performances (as shown in Fig.~\SFpef{}) on face pair matching tasks---accuracy of identifying pairs of different pictures from the same person and rejecting those from different persons---against the Labeled Faces in the Wild (LFW/LFW-a) dataset \cite{LFWTech, wolf2011effective}. Stimulus dimensionalities of the shallow and deep neurons {are} $N=121$ and $N=441$ (i.e.~$11\times11$ and $21\times21$ receptive field sizes) respectively.

\subsection*{Shallow vs.~Deep Representations}

\begin{FPfigure} 
\begin{minipage}{\textwidth} \centering \includegraphics[width=0.9\textwidth]{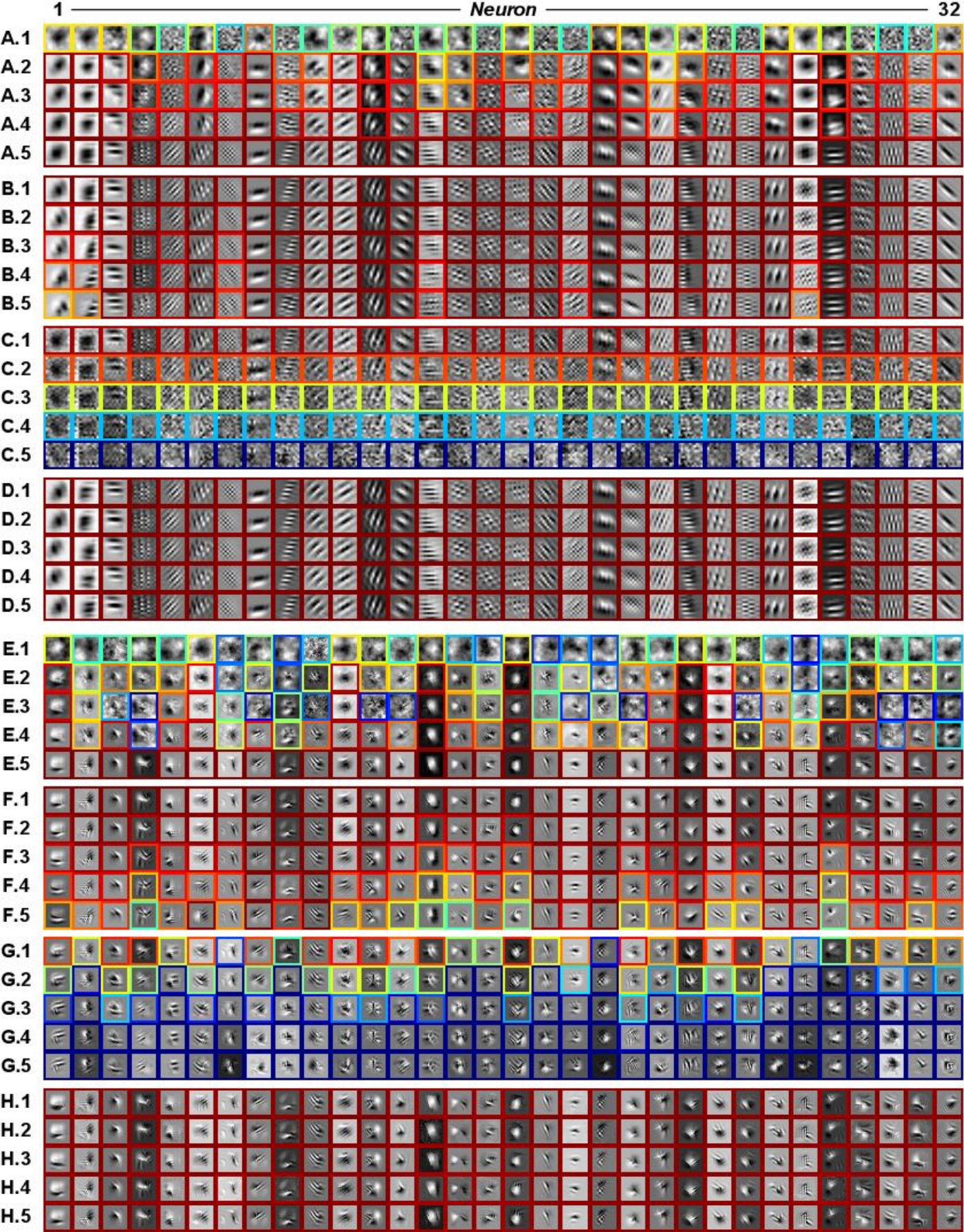} \end{minipage}
\caption{
{\bf Visualization of shallow and deep representations.} (A) Optimal stimulus search trajectories of shallow neurons. Each column of (A) demonstrates the optimal stimulus search trajectory of one neuron, by showing the initial ${1}/{f}$ random stimulus in (A.1), resultant optimal stimulus in (A.5), and 3 intermediate stimuli corresponding to the 3 largest second derivatives (i.e.~curvatures) of the fitness history in (A.2--4). (B) Invariance paths of shallow neurons. Each column of (B) demonstrates the invariance path search results of one neuron, starting from corresponding optimal stimulus as shown in (A.5) and moving away with distance constraints from $\delta = 0.1\pi$ to $0.5\pi$ as shown in (B.1--5) accordingly. (C) Selectivity paths of shallow neurons. Definitions follow (B). (D) Invariance subspaces of shallow neurons. Each column of (D) randomly shows 5 results out of 20 runs of invariance path searches at $\delta = 0.1\pi$. (E--H) Optimal stimulus search trajectories, invariance paths, selectivity paths, and invariance subspaces, of deep neurons, respectively. Definitions follow (A--D). Color indicates fitness (i.e.~response) of a neuron (definition of color map follows Fig.~\ref{fig:methods}).}
\label{fig:ind_res}
\end{FPfigure} 

Figure \ref{fig:ind_res}A and \ref{fig:ind_res}E illustrate the trajectories of optimal stimulus searches of 32 top-layer neurons from the best performing shallow network and 32 from the best performing deep network respectively. Worth to note, to further increase the searching speed, the initial stimulus {was} selected from 1000 ${1}/{f^{\alpha}}$ random stimuli with the best fitness, where $\alpha \in \left\lbrace -4,-3,-2,-1,0 \right\rbrace$, without sacrificing the nature of random initialization. From the results, two major differences can be observed. First, the fitness landscapes of deep neurons are significantly more complex than those of shallow neurons, as a large fraction of searching trajectories for deep neurons actually go down then up before reaching the optima, which also suggests this numerical framework can handle non-convexity reasonably well. Second, the optimal stimuli of deep neurons are also relatively more complex (i.e.~consisting of more textural components) than those of shallow neurons (which are more uni-textured or Gabor-like).


\begin{figure}
\centering \includegraphics[width=0.5\textwidth]{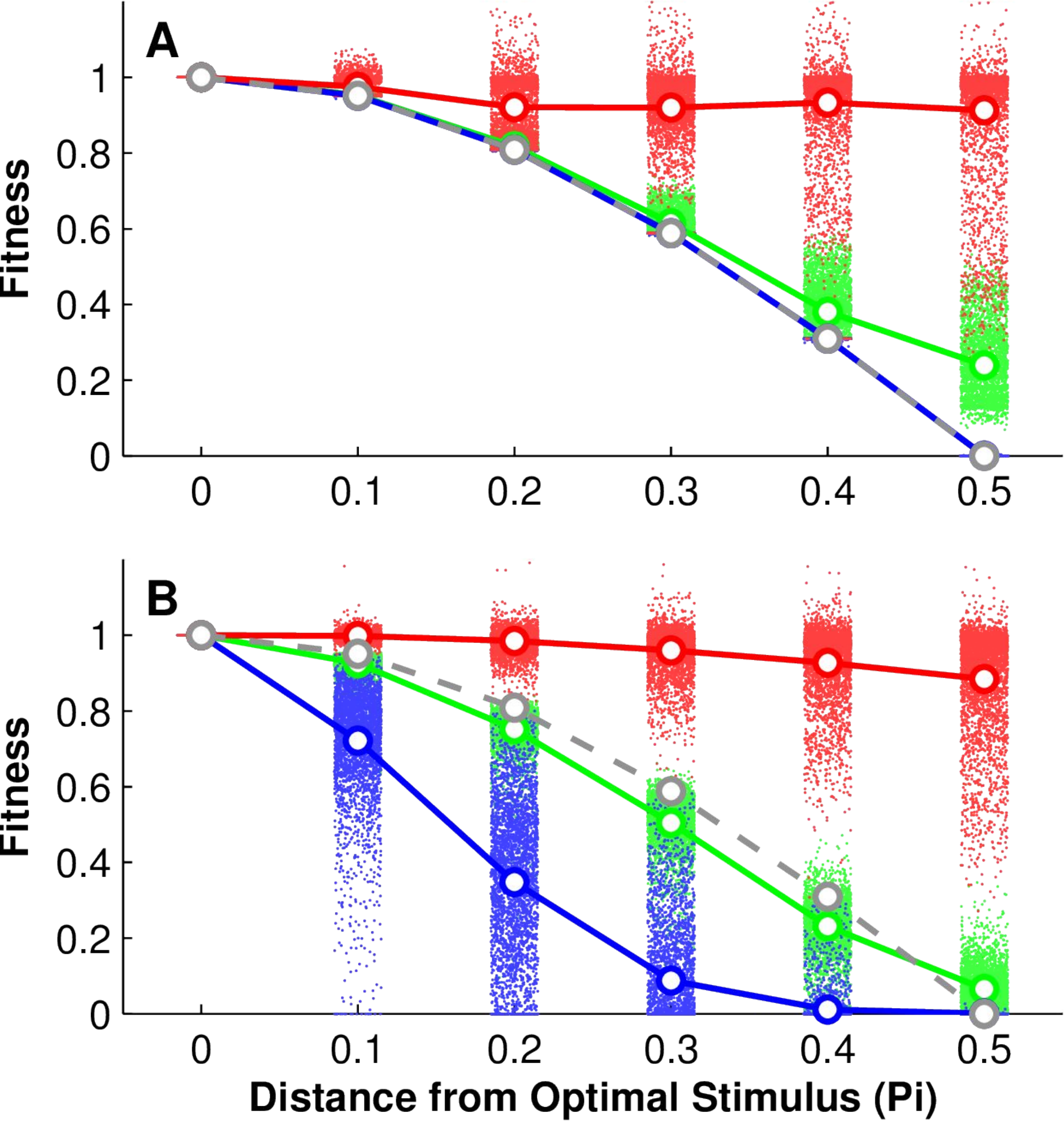}
\caption{
{\bf Comparison of shallow and deep fitness landscapes.} (A) Fitness-distance diagram of shallow neurons. Red, blue and green dots correspond to results of invariance and selectivity path searches, and random walks. Means of results are plotted as solid lines in corresponding colors, and the cosine baseline curve as dashed line in gray. (B) Fitness-distance diagram of deep neurons. Definitions follow (A).}
\label{fig:ind_fdd} 
\end{figure}

Figure \ref{fig:ind_res}B/\ref{fig:ind_res}C and \ref{fig:ind_res}F/\ref{fig:ind_res}G illustrate the invariance/selectivity paths on the same sets of shallow and deep neurons as in Fig.~\ref{fig:ind_res}A and \ref{fig:ind_res}D. It can be observed that, while invariance paths are mostly phase changes and selectivity paths are leading toward meaningless noises (all at the same falloff rate) for shallow neurons, both types of paths consist of sophisticated shape deformations for deep neurons. Also, although most shallow neurons are selective to manually rotated optimal stimuli (i.e.~Gabor filters of different orientations) as well, their fitnesses still do not drop faster then the nonparametric numerical solutions as presented in Fig.~\ref{fig:ind_res}C. In fact, this intriguing difference generalizes across all shallow vs.~deep neurons tested in this work. As the fitness-distance diagrams shown in Fig.~\ref{fig:ind_fdd}, on average, though shallow neurons (Fig.~\ref{fig:ind_fdd}A) have good invariance (i.e.~red curve stays high), they do not have any selectivity (i.e.~blue curve drops slow and completely overlaps with baseline), while deep neurons (Fig.~\ref{fig:ind_fdd}B) show both good invariance and selectivity (i.e.~red curve stays high; blue curve drops fast). Random walks on the fitness landscape \cite{jones1995fitness}, shown as the green curve in Fig.~\ref{fig:ind_fdd}, can be defined as $f\left(\cos\left(\delta\right)\hat{\ib{x}} + \sin\left(\delta\right)\tilde{\ib{x}}\right)$ where $\tilde{\ib{x}}$ is any random stimulus such that $\left\| \tilde{\ib{x}} \right\| = 1$ and $\left\langle \hat{\ib{x}},\tilde{\ib{x}} \right\rangle = 0$, which can be obtained through random projection of $\mathrm{Null}\left(\hat{\ib{x}}\right)$. Comparing the gaps between selectivity and random-walk curves in Fig.~\ref{fig:ind_fdd}A and \ref{fig:ind_fdd}B again shows the selectivity of deep neurons is in fact way more ``selective'' (i.e.~not noise-like) than of shallow neurons. To further characterize the invariance properties, multiple runs of invariance path searches {were} executed and results are visualized in Fig.~\ref{fig:ind_res}D and \ref{fig:ind_res}H. The diversity (i.e.~dimensionality, or subspace capacity) of the ``invariance plateau'' (e.g.~the central high fitness region in Fig.~\ref{fig:methods}) constitution of deep neurons is also more apparent than that of shallow neurons. Worth to note, a small fraction of invariance path search results, as shown in Fig.~\ref{fig:ind_fdd}, in fact have fitnesses higher than the optimal stimuli's due to nature of non-convex problems, which however does not cause noticeable differences in the resultant statistics.

\begin{figure}
\centering \includegraphics[width=0.5\textwidth]{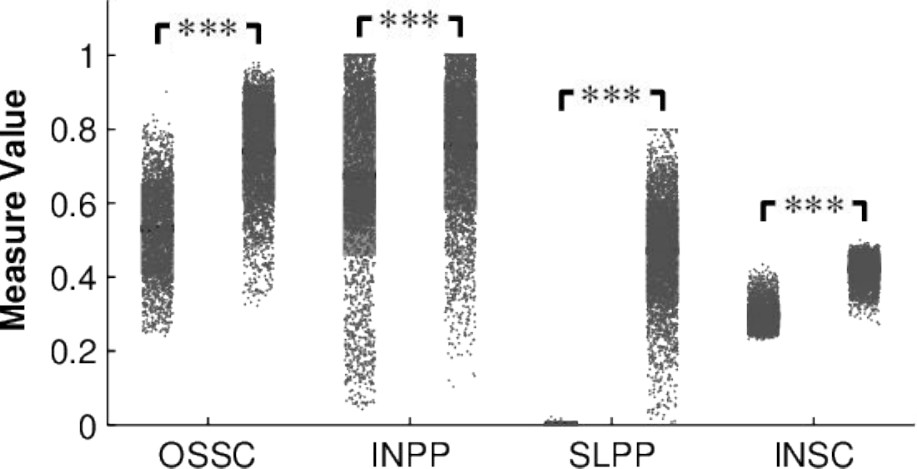}
\caption{
{\bf Comparisons of shallow and deep representation measures.} Representation measures from left to right are, optimal stimulus spectral complexity (OSSC), invariance path potential (INPP), selectivity path potential (SLPP), and invariance subspace capacity (INSC), where data points from shallow networks (left distributions) and deep networks (right distributions) are shown side by side. Significances of differences of means are obtained through permutation tests, and $p < 0.001$ in all four measures. Sensitivity indexes ($d'$) between distributions of shallow and deep representation measures from left to right are 1.63, 0.42, 4.73, and 3.67.} 
\label{fig:ind_mea}
\end{figure}

Comparisons of spectral complexities, invariance/selectivity path potentials, and invariance subspace capacities of shallow and deep neurons (i.e.~as unit representations) are summarized in Fig.~\ref{fig:ind_mea}, where all measures are normalized to have analytical upper and lower bounds being 1 and 0. These measures break down the differences between shallow and deep representations, in addition to the straightforward representation separability/classifiability as used in previous studies \cite{donahue2014decaf, zeiler2014visualizing}. For instance, subtle differences of visual features have better chances being distinguished when the ``gap'' between invariance and selectivity curves (i.e.~the dynamic range of neuronal responses, or amplification ratio of differences) is large. The increase of complexities in deeper neurons' preferred/optimal stimuli also agrees with neurobiological findings in higher visual cortical areas (e.g.~V2 and V4 \cite{hegde2000selectivity, pasupathy2001shape}). In addition, the highly structural representations of neurons from deep networks with purely random convolution kernels found in the experiments explain why such type of random deep networks can still have good performances, on top of the theoretical analysis for shallow networks \cite{saxe2011random}. 


\subsection*{Good vs.~Bad Representations} 

\begin{figure}
\centering \includegraphics[width=0.9\textwidth]{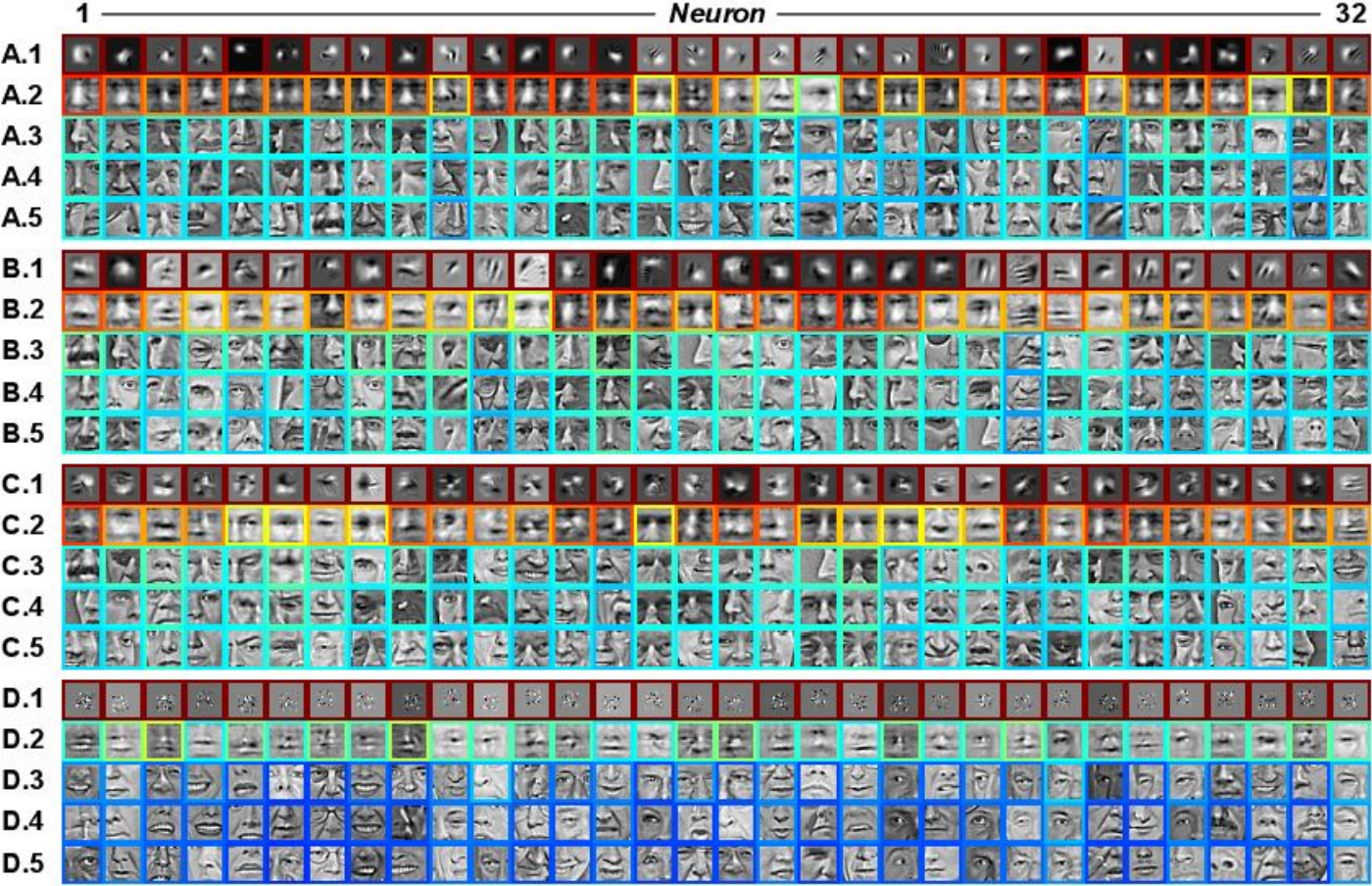}
\caption{
{\bf Explanation power of optimal stimulus.} (A) Deep network with the best average explanation power of optimal stimulus. Each column of row (A.1) shows the optimal stimulus of the corresponding neuron. Each column of row (A.3--5) shows the task-related stimuli with the first, second, and third highest explainabilities (i.e.~inner-product distances to the optimal stimulus) respectively. Each column of row (A.2) shows the average of task-related stimuli with the top 1\% explainabilities. (B--D) Deep networks with the second best, third best, and worst average explanation power of optimal stimulus, respectively. Definitions of rows follow (A). Color indicates explainability, and higher means better (definition of color map follows Fig.~\ref{fig:methods}).}
\label{fig:ind_exp}
\end{figure}


To test how well optimal stimulus may linearly explain task-related stimuli and how the explanation power correlates with network's performance, 10,000 image patches {were} randomly sampled out of the predefined facial regions of pictures from the LFW-a dataset \cite{wolf2011effective} and compared against the optimal stimuli of all 32 top-layer neurons from all 100 deep networks. Figure \ref{fig:ind_exp} shows the results of best 3 (in \ref{fig:ind_exp}A--C) and worst 1 (in \ref{fig:ind_exp}D) networks in terms of explanation power. For simplicity and clarity of presentation, the explanation power measure is calculated using the top 1\% (i.e.~100) of the task-related stimuli with highest explainabilities, without biasing the results compared to using all task-related stimuli. It can be observed that, though none of the task-related stimuli is highly similar to any of the optimal stimuli, there is an $R^2=0.36$ correlation (with details shown in Fig.~\SFexp{}) between network's explanation power and performance; however, when alternatively comparing the averages of the top 1\% of task-related stimuli and the optimal stimuli, the similarities (Fig.~\ref{fig:ind_exp}A.1 vs.~\ref{fig:ind_exp}A.2, etc.) and dissimilarities (Fig.~\ref{fig:ind_exp}D.1 vs.~\ref{fig:ind_exp}D.2) become apparent. 


\begin{figure}
\centering \includegraphics[width=0.5\textwidth]{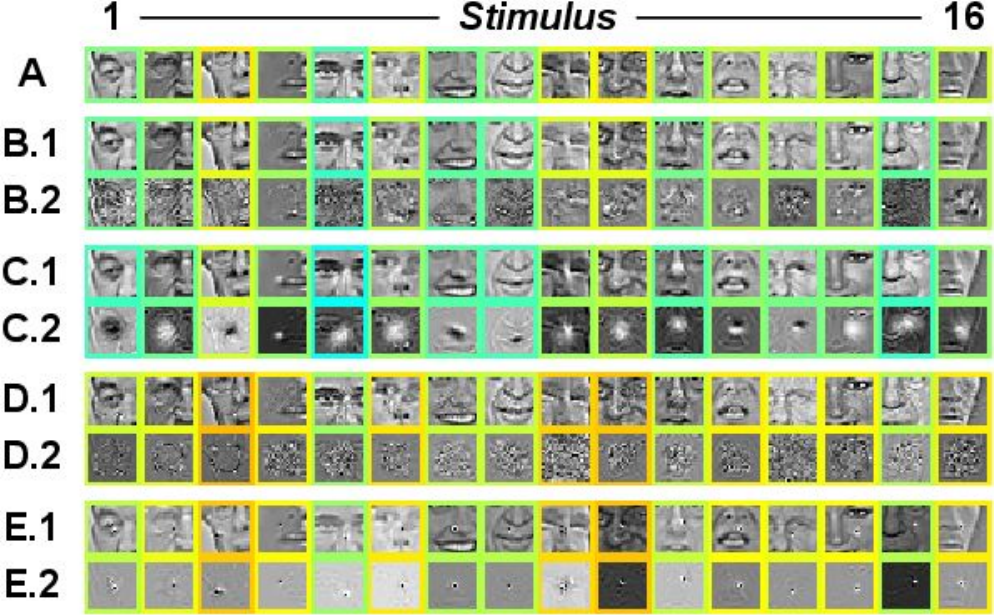}
\caption{
{\bf Subspace alignment between invariance and selectivity paths and task-related stimuli.} (A) Reference stimuli used as starting points of invariance and selectivity path searches. (B) Best invariance subspace alignments. Each column of row (B.1) shows an invariance path search result with the best alignments against the principle component vector space of task-related stimuli (i.e.~the eigenface vector space). Each column of row (B.2) shows the difference between result in (B.1) and corresponding reference stimulus in (A) for better visual comparison. All columns do not necessary come from the same deep network. (C--E) Best selectivity subspace alignments, worst invariance subspace alignments, and worst selectivity subspace alignments, respectively. Definitions of rows follow (B). Color indicates alignment measure, and lower means better (definition of color map follows Fig.~\ref{fig:methods}).} 
\label{fig:pop_aln}
\end{figure}

It has been shown that invariance and selectivity are extremely important for recognition in various modalities and species \cite{desimone1991face, ito1995size, quiroga2005invariant}; however, how these properties directly affect the recognition performances remains unclear. To address this puzzle, starting from 16 randomly sampled reference stimuli (as shown in Fig.~\ref{fig:pop_aln}A), invariance and selectivity path searches {were} performed at $\delta = 0.1\pi$ using Eq.~(\ref{eq:I2}, \ref{eq:S2}) for population representations. Figure \ref{fig:pop_aln}B--E demonstrates the results of invariance and selectivity subspace alignment analyses. It can be observed that, good alignments (i.e.~sparser representations in the principle component vector spaces) correspond to highly structural deformations or lighting changes, which are known to be important factors in visual recognition, while bad alignments correspond to mostly meaningless noises. The correlation between network's invariance/selectivity subspace alignment measure and performance is $R^2=0.19/0.31$ (with details shown in Fig.~\SFaln{}). Network's invariance subspace capacity (in terms of population representation), which though has an $R^2=0.15$ correlation as well, is in fact negatively correlated to network's performance (with details shown in Fig.~\SFinc{}), mainly due to the fact that poor performing network's invariance path search results are usually noisier (as depicted in Fig.~\ref{fig:pop_aln}D), and thus should be interpreted differently from the unit representation cases. Invariance/selectivity path potentials for population representations, on the other hand, are weaker in explaining networks' performances (with $R^2=0.10/0.06$), due to the fact that all deep networks have high but less ranked invariance and selectivity (as the fitness-distance diagram shown in Fig.~\SFfda{}).

\begin{figure}
\centering \includegraphics[width=0.5\textwidth]{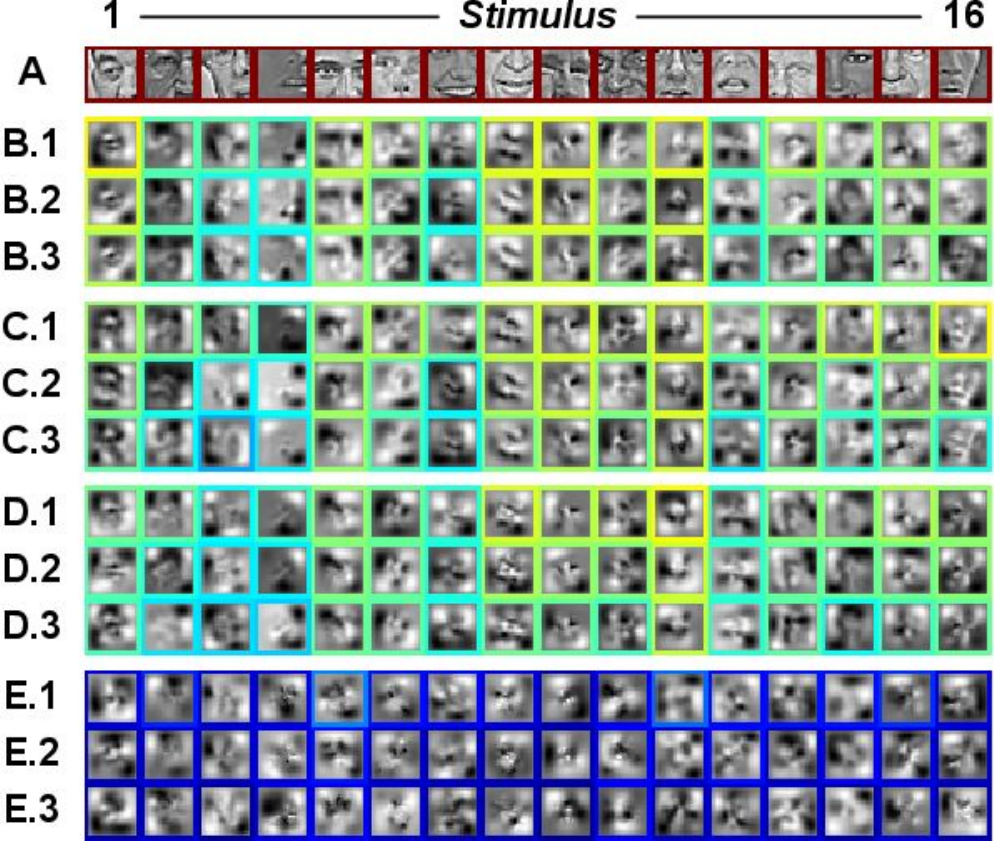}
\caption{
{\bf Encoding specificity of task-related stimulus.} (A) Reference stimuli used as sources of representations. (B) Best encoding specificities. Each column of row (B.1--3) shows 3 reconstructions of the corresponding reference stimulus in (A) with the best SSIM measures (out of 10 reconstructions), from deep network of the best average encoding specificity. All columns do not necessary come from the same deep network. (C--E) Second best, third best, and worst encoding specificities, respectively. Definitions of rows follow (B). Color indicates SSIM measure, and higher means better (definition of color map follows Fig.~\ref{fig:methods}).}
\label{fig:pop_ens}
\end{figure}

To further understand the meaning of population representation $\ib{r}$ formed by a deep network $f$, the inverse function $f^{-1}\left( \ib{r} \right)$ {was} approximated through the numerical optimization framework to reveal what kind of stimulus can drive the deep network to give output $\ib{r}$. Instead of inverting randomly generated $\ib{r}$, which may not have feasible or interpretable solutions, we {used} $\ib{r}$ from known reference stimulus $\ib{x}^{*}$ to study the encoding specificity and its relationship to performance. Figure \ref{fig:pop_ens} shows the reference stimuli (in \ref{fig:pop_ens}A), the best 3 (in \ref{fig:pop_ens}B--D) and worst 1 (in \ref{fig:pop_ens}E) examples of encoding specificities, and the overall correlation between encoding specificity and performance is $R^2=0.41$ (with details shown in Fig.~\SFenc{}; SSIM measured using default settings of the standard implementation \cite{wang2004image} while different settings also yield similar correlations; other similarity measures like SNR, PSNR, and inner-product distance yield similar correlations as well). It can be observed that, though the reconstructed stimuli are overall of lower spatial frequencies, SSIM is still able to capture the relative similarities (e.g.~reconstructions of task-related stimulus 1 and 5).

\begin{figure}
\centering \includegraphics[width=0.5\textwidth]{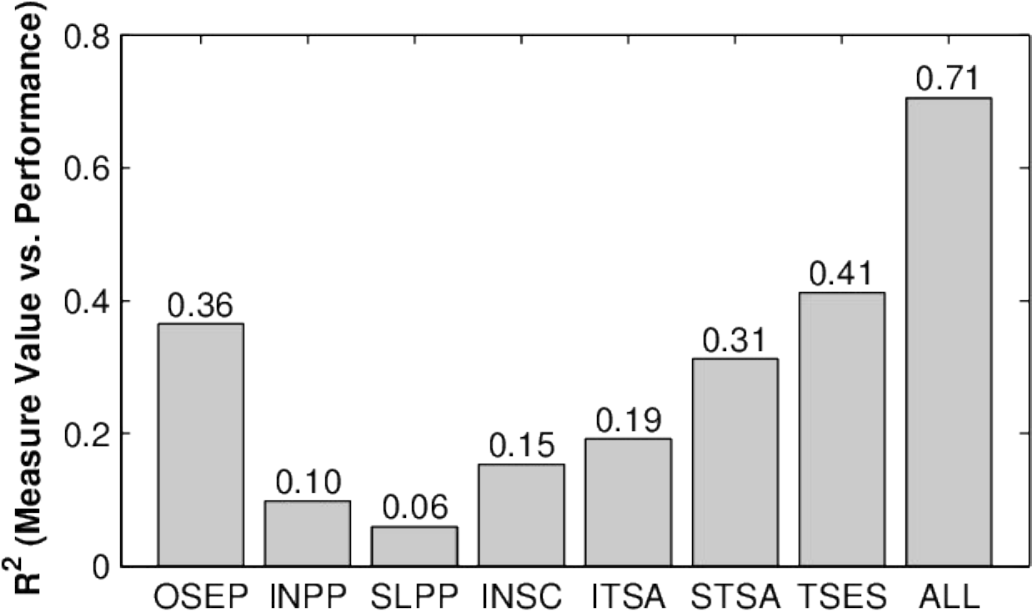}
\caption{
{\bf Spearman's correlations between deep network's representation measures and performance.} Representation measures from left to right are, optimal stimulus explanation power (OSEP), invariance path potential (INPP), selectivity path potential (SLPP), invariance subspace capacity (INSC), invariance vs.~task-related stimuli subspace alignment (ITSA), selectivity vs.~task-related stimuli subspace alignment (STSA), task-related stimulus encoding specificity (TSES), and all measures together (ALL) using linear multiple correlation analysis. Pearson's correlations following the same order are 0.37, 0.02, 0.02, 0.19, 0.17, 0.23, 0.39, and 0.69.}
\label{fig:all_R^2}
\end{figure}


Spearman's correlations between all representation measures and deep networks' performances are summarized in Fig.~\ref{fig:all_R^2}, and 71\% of the variance of deep networks' performances can be explained by the proposed measures altogether. These measures further break down the differences between deep representations and suggest features like network's explanation power, invariance and selectivity subspace properties, and encoding specificity against the task-related stimuli can be extremely important for the effectiveness of representations. Encoding specificity, as a single representation measure, can best explain the deep network's performance. Though bearing strong similarity to the invariance path search in terms of mathematical formulations, encoding specificity is in fact a radically different and complementary measure, since its numerical optimization is not constrained through distances to the reference stimulus (while distance constraints can induce \emph{a priori} similarities); thus, it can better function as an unconstrained global characterization of the ``encoding landscape'' (i.e.~multivariate tuning landscape of a population) and estimate if the (reconstructed) stimuli that a deep network is invariant to are ``selective'' (i.e.~visually similar to original stimuli). Explanation power of optimal stimulus is the second best single representation measure. Following the efficient coding theory \cite{barlow1961possible, laughlin1981simple}, one may readily exclude the possibility of an optimal stimulus being highly similar to the task-related stimuli. However, in a high-dimensional stimulus space, how to optimally construct the tuning landscape (and place the optimal stimuli) remains unclear, and whether the observed phenomenon---optimal stimuli of well performing networks being highly similar to the averages of neighboring task-related stimuli (as shown in Fig.~\ref{fig:ind_exp})---is theoretically essential to an optimal tuning landscape still needs to be further studied, even though similar phenomenon was also shown in previous studies \cite{le2012building, simonyan2013deep}. Such kind of ``linearities'' around the task-related stimuli out of these highly nonlinear networks may also explain why natural stimuli can be more efficient in estimating the tuning of biological neurons \cite{talebi2012natural}. 


\section*{Discussion} 
\label{sec:discussion}

\subsection*{Why Do We Need a New Framework?} 

As briefly reviewed in \nameref{sec:intro}, the study of sensory representations within biological neural networks has been an extremely important topic for decades and numerous methods have been proposed; nevertheless, most methods can be viewed as instances of certain generalized functional frameworks \cite{wu2006complete, dimattina2013adaptive}.
We argue than existing frameworks being actively used are still mostly highly parametric, in terms of both the model assumption and stimulus generation.
When moving away from peripheral or primary sensory circuitry and going deeper into higher-level sensory cortices, the modeling power of parametric models becomes inevitably insufficient, and even though parametric stimulus may facilitate the speed of representation characterization, its inherently assumptive nature can strongly bias the results.
Although methods utilizing nonparametric artificial neural networks to model sensory circuitry also have been long known \cite{lau2002computational, prenger2004nonlinear}, interpreting the resultant models remains difficult \cite{wu2006complete}.
This directly links to the fact that methods for characterizing/interpreting artificial neural networks themselves are still relatively oversimplified and scarce, and sometimes algorithms using handcrafted representations (e.g.~HoG and SIFT \cite{szeliski2010computer}) are still preferred for their interpretability.
Such problems only get worse when deeper networks are being adopted (either as system identification models, or as recognition algorithms), since understanding them also becomes even harder.
Another important issue is the lack of principled and efficient approach for characterizing neuronal populations, due to the fact that, e.g., there is no straightforward generalization of the definition of optimal stimulus to neural populations, and existing methods can suffer from the exponential growth of required measurements with respect to the population size \cite{dimattina2013adaptive}.


Major benefits of the proposed framework can be categorized as follows.
(1) Generality: the proposed framework tackles the challenges in studying sensory representation in a highly generalizable and unbiased way via supporting both nonparametric models and nonparametric stimuli.
Although not explicitly tested in this work, such setting in practice can be easily reconfigured and extended to support both parametric models, by numerically analyzing them as nonparametric models (especially when analytical solutions are not easily derivable), and parametric stimuli, by simply reframing the optimization in the parameter space $\ib{p} \in \mathbb{R}^P$ and evaluating the fitness through the generative function $f\left(\ib{x}\left(\ib{p}\right)\right)$ (especially when certain stimulus type or deformation is confirmed to be highly relevant).
As described in \nameref{sec:methods}, the proposed framework also supports characterizing both unit and population representations, and the fitness-distance diagram can be viewed as a ``generalized tuning curve'' characterization of the landscape, which is dimensionality- and population size-insensitive as well.
(2) Efficiency: the proposed framework is designed to be efficient using low-rank and stochastic approximations of the complex high-dimensional tuning landscapes. 
As reported in multiple existing works \cite{touryan2002isolation, rust2004spike} where second-order models {were} utilized for system identification and representation characterization, often only a few number of eigenvectors corresponding to the most positive and negative eigenvalues have significant and interpretable spatial structures.
Directly searching for invariance and selectivity paths eases this problem and reduces potentially costly measurement requirement, which is on the order of $\mathcal{O}\left(N^2\right)$, to $\mathcal{O}\left(N\right)$, in addition to the fact that non-local characterization can be obtained in this work as well.
Stochastic sampling of the solution space of the invariance subspace capacity provides an efficient approximation of the intrinsic dimensionality estimation approach \cite{kegl2002intrinsic, levina2004maximum}, and, of the encoding specificity, an efficient alternative of the maximally informative ensemble estimation approach \cite{machens2002adaptive}, both of which can require substantially more measurements in order to yield robust and meaningful results.
As also mentioned in \nameref{sec:methods}, the back-end solver with relatively simple constraints positively contributes to the speed of this framework as well. 

\subsection*{How Close Are We to Fully Understanding Deep Networks?} 

Via the proposed numerical optimization framework, the following important aspects of sensory representation research are enabled or facilitated: (1) Decoding unit or population representation (as optimal or reconstructed stimulus). (2) Discovering what changes in stimulus the representation is invariant or selective to. (3) Explaining how deep network's representation may affect its task performances. Main findings of this work can be summarized as:
\begin{itemize} 
\item Complexity of representation increases along network's depth.
\item Unlike deep representation which is both invariant and selective, shallow representation is only invariant, not selective, and its capacity of invariance is significantly lower than deep representation's.
\item How well the optimal stimulus can ``explain'' task-related stimuli, and how specific the representation encoding of task-related stimuli is, both decently explain network's task performance.
\item How well the invariance and selectivity of representation align with the actual ``distribution'' of task-related stimuli also partially explains network's task performance.
\end{itemize}
We argue the importance of these findings as follows.
First, the experiments {were} conducted in a large number of neurons and all results are statistically significant.
Second, the results {were} obtained with relatively unbiased methods (i.e.~nonparametric model plus nonparametric stimulus).
Third, the resultant representation measures match what are known to be crucial properties for performing visual recognition accurately, and explain network's performance decently.
Finally, the results also assure the viability and effectiveness of integrating extrinsic properties (i.e.~measures based on task-related stimuli) into deep network characterization, in addition to only using intrinsic properties (i.e.~measures based on optimal stimuli).
In fact, if just using the four intrinsic properties (i.e. OSSC, INPP, SLPP and INSC as shown in Fig.~\ref{fig:ind_mea}; not included in Fig.~\ref{fig:all_R^2}) of the unit representations within deep networks (with measures averaged from all 32 top-layer neurons), we can only explain the network's performance with $R^2 = 0.34$, which is significantly lower than the correlation based on extrinsic properties (i.e.~$R^2 = 0.71$; combining both properties only marginally increases $R^2$ to 0.73).
This also corresponds to our observation that the ranking of deep networks may differ from task to task (e.g.~face pair matching in this work vs.~object recognition in \cite{pinto2009high}), and the fact that neural circuitry can (and maybe constantly) shape its tuning (i.e.~representation) dynamically according to the tasks being performed \cite{gilbert2007brain}, such that ``intrinsic'' properties may not mean much.
Also, combining the results from machine learning and computer vision studies on deep network's superior theoretical efficiencies \cite{delalleau2011shallow, montufar2014number} and actual performances over shallow network's, we argue that network depth is possibly the most plausible way of implementing efficient sensory processing.


However, there are still undeniably plenty of unsolved and even unaddressed mysteries about deep networks.
Here we list the ones we think to be of the most importance.
First, even though we have successfully visualized the representations of certain deep neurons, like what is argued in \nameref{sec:results} about the optimal construction of tuning landscapes, how these representations explicitly facilitate generalization (i.e.~invariance) across features that are distinct but of the same class, and discrimination (i.e.~selectivity) between features that are similar but of different classes, remains ambiguous.
Second, even though some smaller-scale studies with heuristic evaluation \cite{zeiler2014visualizing} or theoretical characterization \cite{saxe2013exact} of representations' evolution have been conducted, how the training algorithms (e.g.~backpropagation) explicitly (and orientedly) shape the representations, and how different training settings alter the resultant representations, remain largely unclear.
Third, the ability of ``bottom-up attention'' seems to be prominently functioning in these feedforward deep networks \cite{zeiler2014visualizing, simonyan2013deep} for unaligned and mildly cluttered visual object recognition. Even though pooling operations are generally considered capable of providing such function, to what degree it (or other operations) contributes to this, how it explicitly (or implicitly) resolves, e.g., objects' labels from cluttered scenes, and how much variation can be tolerated, remain unclear.
Finally, some ``higher-level vision like'' tasks also have been reported performing decently by deep networks (e.g.~image style recognition \cite{karayev2013recognizing} and scene understanding \cite{CVPR14_Khosla}), where the representations being used can be even more ambiguous and undefined, and more ``holistic'' instead of part-based representations are likely to dominate. In such scenarios, object recognition alone can become an overly limited definition of vision, and may not best serve as the framing problem for studying high-level vision and its representations \cite{cox2014we}.
To sum up, with all these unexplained properties of deep networks, how to precisely address these mysteries, arguably would be the most important and fruitful question to be kept asking.

\subsection*{Challenges and Possible Directions}

In addition to the unexplained properties listed above, which still lack clearer definitions and directions, here in the following, with no specific order, we also give a list of more well defined challenges that need to be solved, and possible directions to tackle these challenges.
Arguably, sensory representations of the most interest are at neither the bottom layer nor the top layer, but intermediate layers, since the former are relatively well understood and usually consist of Gabor (or Gabor-like) filters, and the later are highly class selective, as directly suggested by the ``behavioral'' (i.e.~classification) results and the visualizations provided in \cite{donahue2014decaf, azizpour2014generic}.
However, for the most advanced deep networks \cite{szegedy2014going, simonyan2014very} which can have around 20 layers of operations, even intermediate layers can be very computationally expensive to measure and may have way more complex tuning landscapes (compared to the two-level deep networks experimented in this work, which consist of up to 8 layers of operations). 
Instead of CPU-based deep network simulation adopted in this work, GPU- or FPGA-/ASIC-based simulation may be required to tackle this potential problem of running speed.
The back-end numerical solver used in this work may not either suffice the needs for even deeper networks; nevertheless, the proposed framework itself is neutral to the selection of solvers, thus more advanced numerical optimization algorithms can always be adopted to improve the quality of numerical solutions.
Similar challenges may be encountered when moving from purely feedforward architectures into more biologically plausible ones---networks with feedback pathways, or recurrent networks---for their widely known nature of complex dynamics.
Though in this work, around 70\% of the variance of performance differences on face pair matching can be explained with the proposed representation measures, whether this number can be further improved, and if these measures can generalize well enough onto even deeper representations, or radically different types of tasks (e.g.~phoneme recognition), remain unclear.
Therefore, the validation of existing, and design of new representation measures, shall continue to be an important direction for future researches. 


Another difficult yet potentially fruitful future challenge is to apply this framework to real biological deep networks, as we suggested in the very beginning of this paper.
While results are shown exclusively for artificial networks, the methods presented in this work by nature require a smaller budget of neuronal measurements that they could potentially be applied to real neurons in real experiments, even when spiking variability is taken into account, given that at least two effective techniques---stochastic averaging and measurement reweighting---are built in the core of CMA-ES, the back-end solver, for robustness against measurement noises.
In fact, we argue that the mathematical formulation of CMA-ES can be viewed as an extension/generalization of the spike-trigger covariance method, which is known to be working effectively for biological neurons \cite{rust2004spike}.
Potential inefficiency in measurement arising from neural adaptation \cite{kohn2007visual} can also benefit from, e.g., Particle Swarm CMA-ES \cite{muller2009particle} or other multipopulation/multimodal optimization methods, where evaluation of stimuli from different local searches can be interwoven to minimize neural adaption and maintain the search speed.
Overall, since the proposed framework supports efficient nonparametric search and analysis, a biological neural network can be characterized \emph{as is} to reduce the chances of obtaining premature or biased numerical results.
On the other hand, based upon recent studies on deep network's capability of encoding representations similar to mammalian visual systems' \cite{yamins2014performance, cadieu2014deep, khaligh2014deep}, we argue that deep networks can also potentially serve as a good surrogate model for understanding the biologically plausible ways of representation encoding.
In addition to the tools proposed in this work, neurophysiological methods like patch clamping, lesioning, ``gene'' (i.e.~hyperparameter) knockout/knockdown, etc.~may as well be used to inspire the development of new tools for characterizing artificial deep networks.
For instance, the effect of surround suppression {was} also observed in the experimented artificial neurons as visualized in Fig.~\ref{fig:ind_res} (i.e.~flat gray peripheries in the optimal stimulus patterns; responses {dropped} when manually {placed}, e.g., dots in peripheries) and, in our preliminary tests, {could} be easily shutdown via ``lesioning'' the prenormalization operations (i.e.~early-stage lateral inhibition) in the network, which is consistent with neurophysiological findings \cite{cavanaugh2002selectivity}.


Finally, though via the proposed numerical framework, we have demonstrated how previously unknown or unconfirmed properties can be identified with high statistical significances, we note this is not equivalent to what will be needed to ``fully'' explain the observed properties and ultimately everything about sensory representations within deep networks---a theoretical framework yet to be constructed/completed.
Notable recent works on theorizing deep convolutional networks include magic theory \cite{anselmi2013unsupervised}, scattering transform \cite{mallat2012group}, convolutional kernel network \cite{MairalKHS14}, etc.
Saxe et al.~\cite{saxe2011random} and Szegedy et al.~\cite{szegedy2013intriguing} also addressed particular phenomena observed in deep networks using theoretical approaches, in addition to the theoretical studies on deep network's properties \cite{delalleau2011shallow, montufar2014number} mentioned earlier.
While undoubtedly different theories should be actively researched and interactively inspired by each other, we also argue that the proposed numerical framework can potentially (and maybe crucially) function as an independent ``diagnostic tool'' for examining, verifying, and facilitating the construction of theories.






\section*{Supporting Information}

\newcommand{\beginsupplement}{\setcounter{figure}{0} \renewcommand{\thefigure}{S\arabic{figure}}}
\beginsupplement

\begin{figure}[H]
\centering \includegraphics[width=0.6\textwidth]{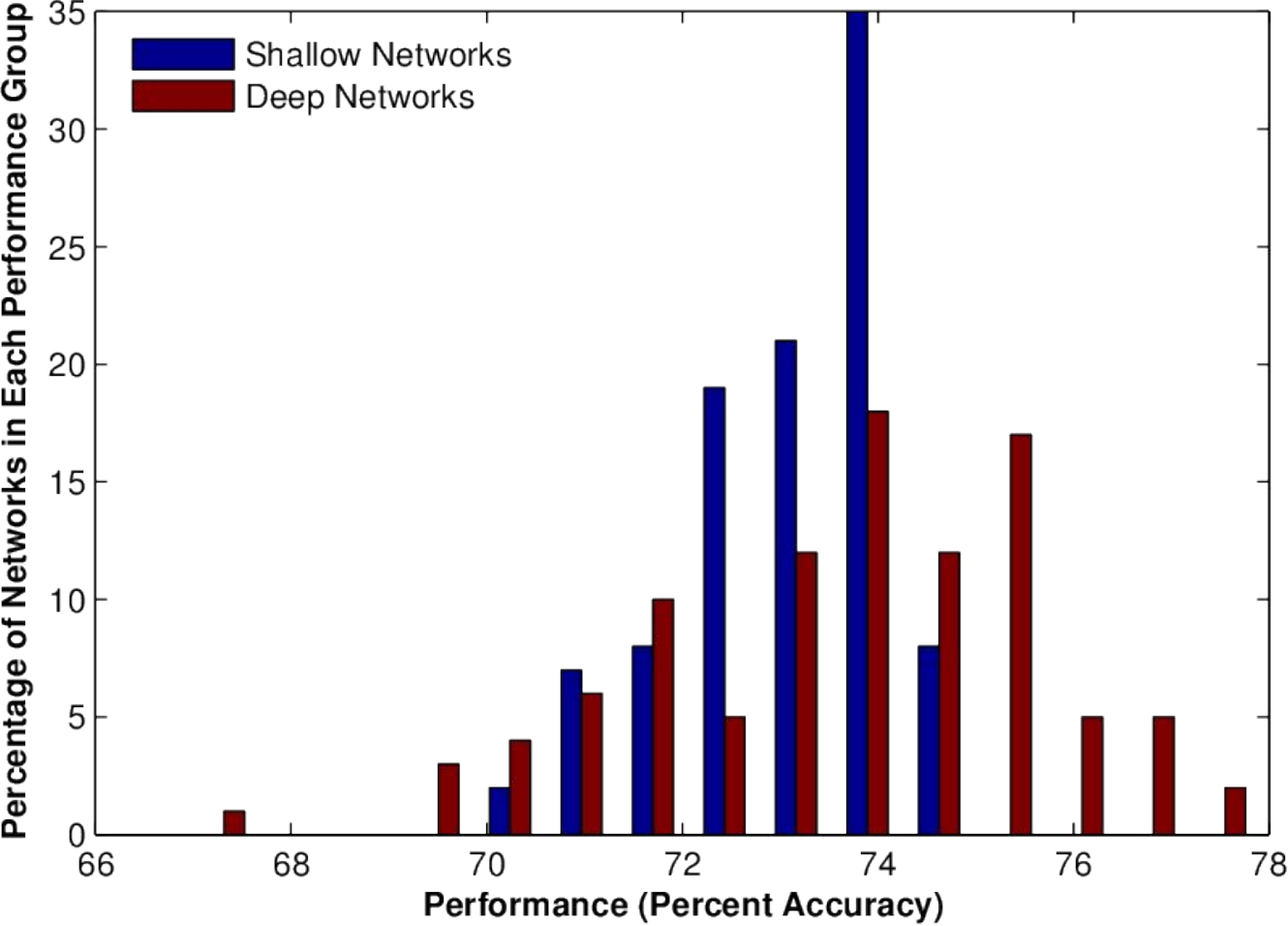} 
\caption{ 
{\bf Distributions of shallow and deep network performances.} Performances of the randomly generated 100 shallow and 100 deep networks {were} obtained via 10-fold cross validations on 6,000 face pairs. Feature vectors used in the linear SVM classifier {were} generated via sliding the receptive fields throughout entire images and concatenating the population representations. Maximum and mean of performances of the deep networks are both significantly higher than those of the shallow networks (under permutation tests, $p<0.001$ and $p=0.008$ respectively). See \cite{cox2011beyond} for more details.}
\label{fig:SFpef}
\end{figure}


\begin{figure}[H]
\centering \includegraphics[width=0.6\textwidth]{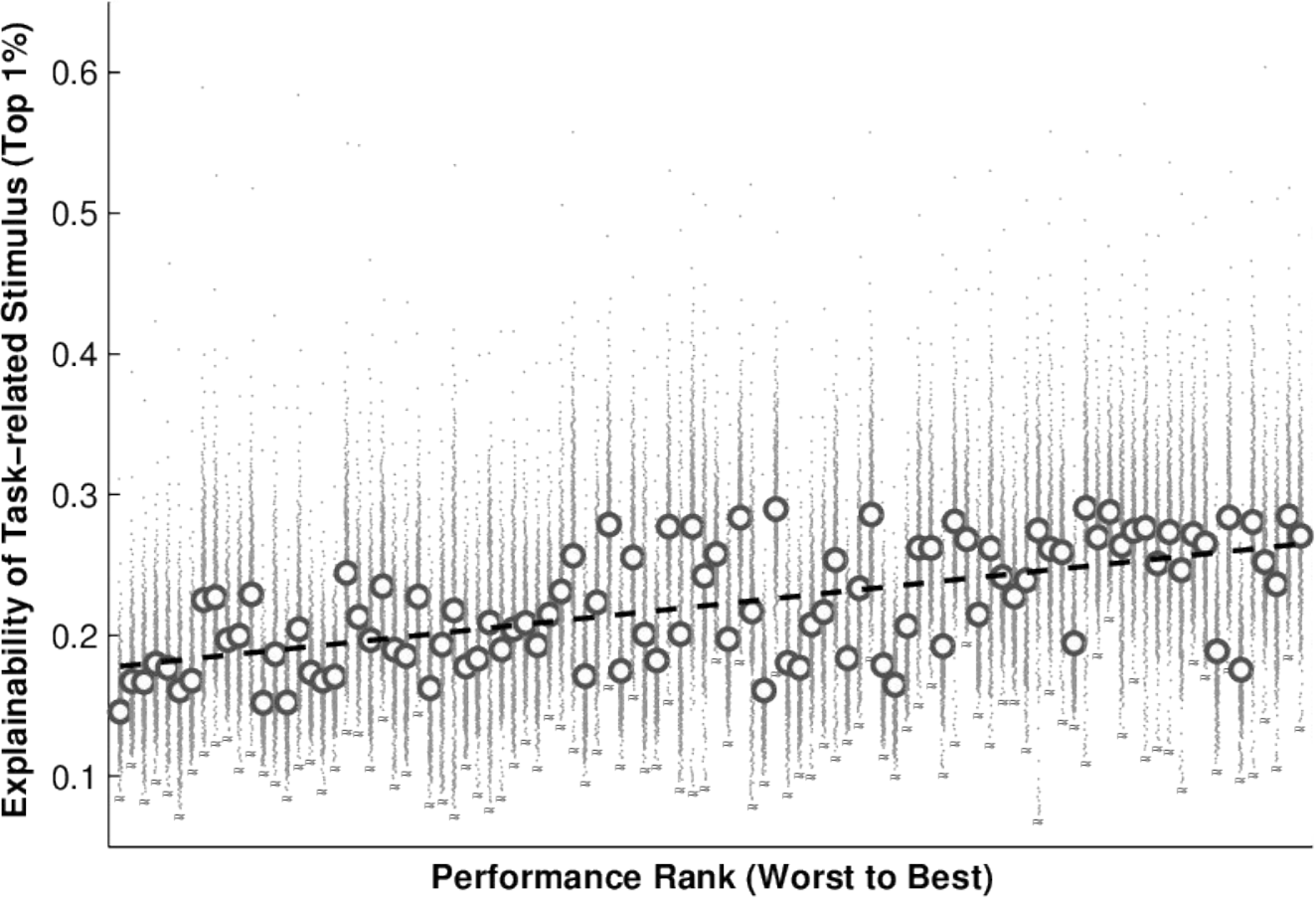} 
\caption{ 
{\bf Explanation power vs.~performances of deep networks.} Each distribution from left to right shows the top 1\% explainabilities of the task-related stimuli against all of the 32 top-layer unit representations of a deep network. Performance ranks, instead of performance values, are used for visualization purposes. Means of distributions are plotted as gray circles and the linear regression as dashed black line. Significance of slope of means under permutation test has $p < 0.001$.}
\label{fig:SFexp}
\end{figure}

\begin{figure}[H]
\centering \includegraphics[width=0.6\textwidth]{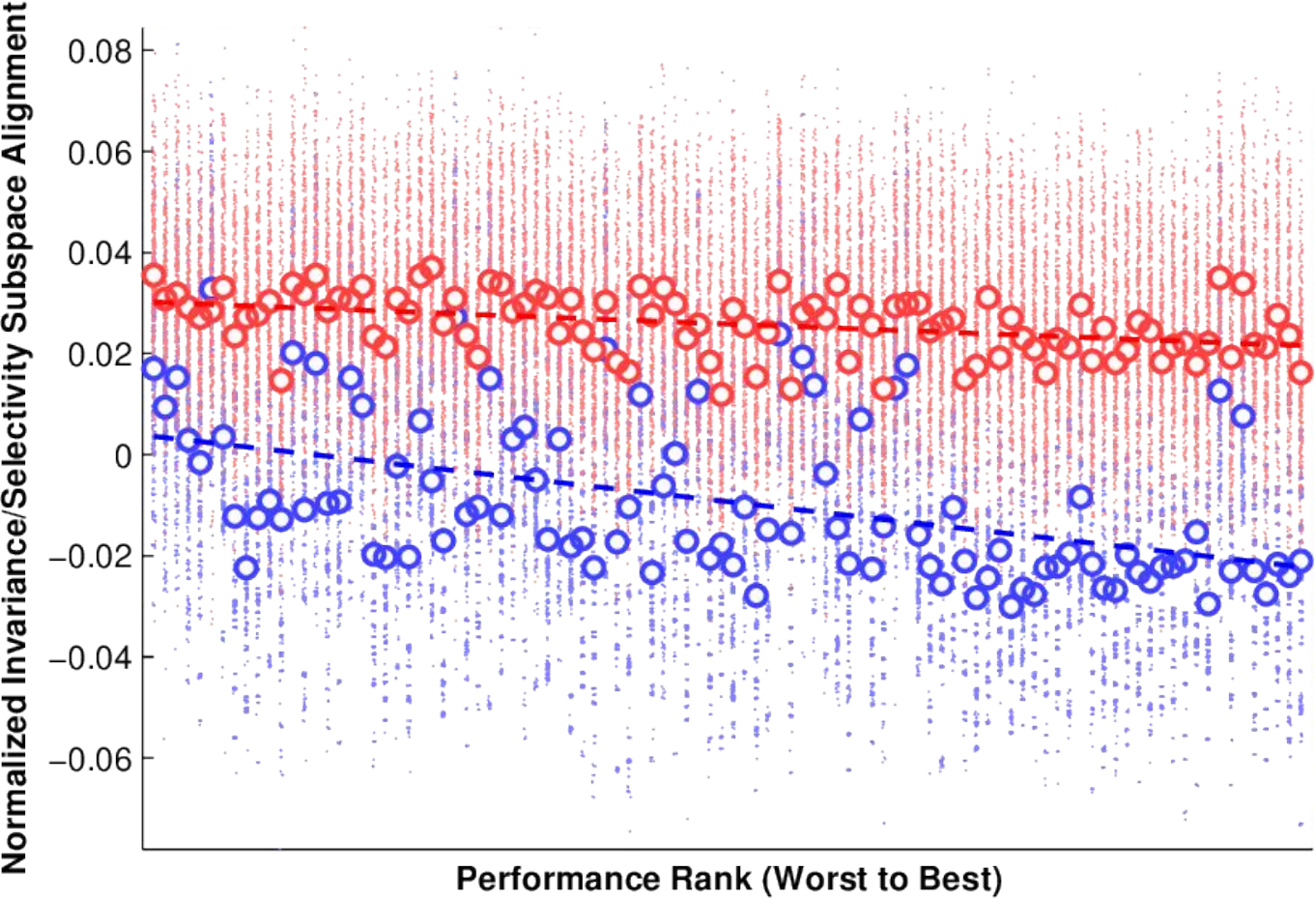}
\caption{ 
{\bf Invariance and selectivity subspace alignments vs.~performances of deep networks.} Each red distribution from left to right shows the invariance subspace alignments against the task-related stimuli in population representations of a deep network given the 16 reference stimuli. Alignment scores are normalized by subtracting the corresponding reference stimuli's own alignment scores. Performance ranks, instead of performance values, are used for visualization purposes. Means of distributions are plotted as red circles and the linear regression as dashed red line. Blue distributions for selectivity subspace alignments follow the same definitions. Significances of slopes of means under permutation tests both have $p < 0.001$.}
\label{fig:SFaln}
\end{figure}

\begin{figure}[H]
\centering \includegraphics[width=0.6\textwidth]{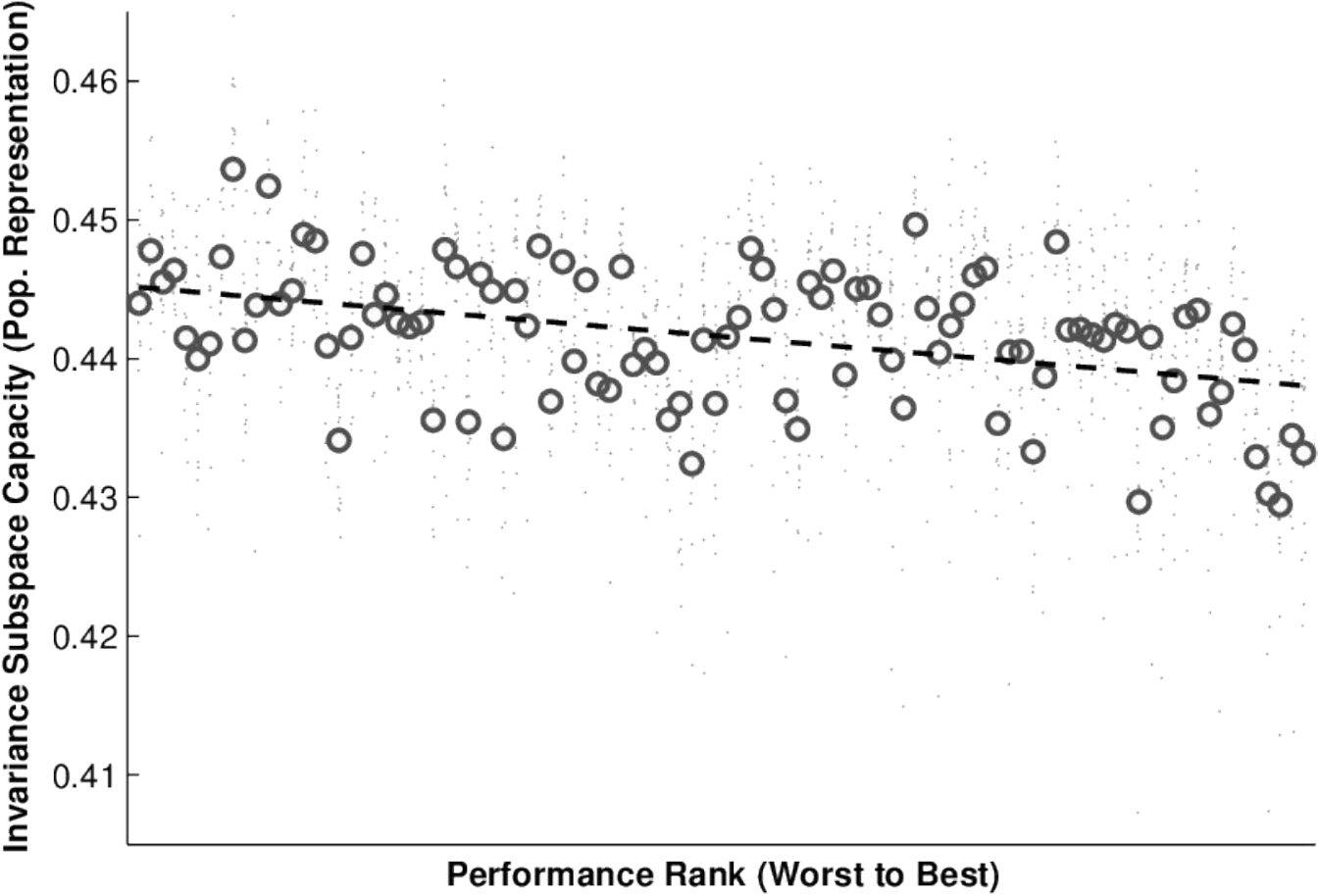}
\caption{ 
{\bf Invariance subspace capacities vs.~performances of deep networks.} Each distribution from left to right shows the invariance subspace capacities of population representations of a deep network given the 16 reference stimuli. Performance ranks, instead of performance values, are used for visualization purposes. Means of distributions are plotted as gray circles and the linear regression as dashed black line. Significance of slope of means under permutation test has $p < 0.001$.}
\label{fig:SFinc}
\end{figure}

\begin{figure}[H]
\centering \includegraphics[width=0.6\textwidth]{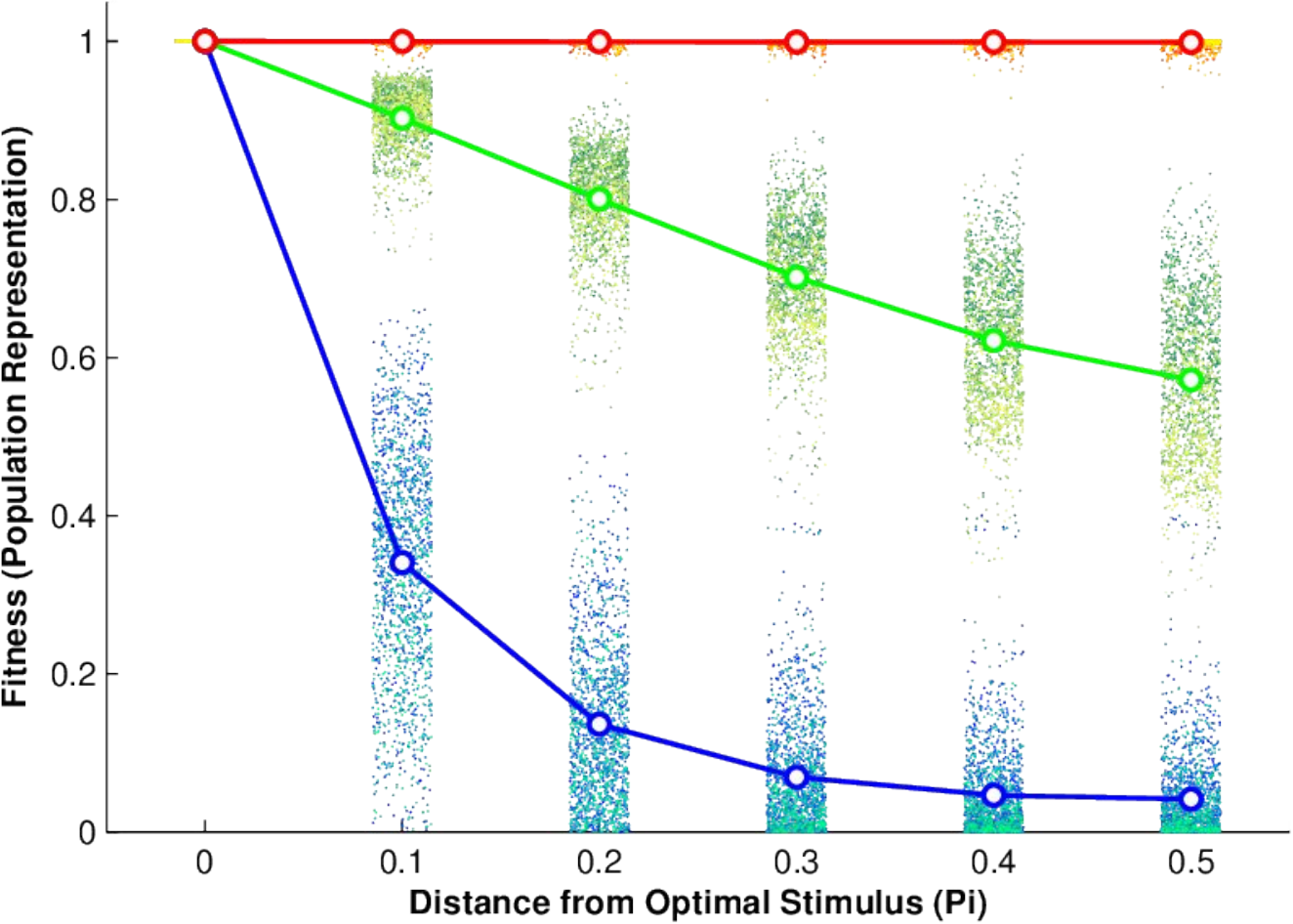}
\caption{ 
{\bf Fitness-distance diagram of population representations within deep networks.} Red, blue, and green dots correspond to results of invariance and selectivity path searches, and random walks, where brighter shades of a color are from better performing networks, and darker shades from poorer performing networks. Means of results are plotted as solid lines in corresponding colors. As visualized, correlations between invariance and selectivity path potentials and performance are weak. Random walk results vs.~performance, on the other hand, has $R^2 = 0.18$ correlation which may arise from the differences in sensitivities to noises, though incorporating it into representation measures does not improve the final multiple correlation.}
\label{fig:SFfda}
\end{figure}

\begin{figure}[H]
\centering \includegraphics[width=0.6\textwidth]{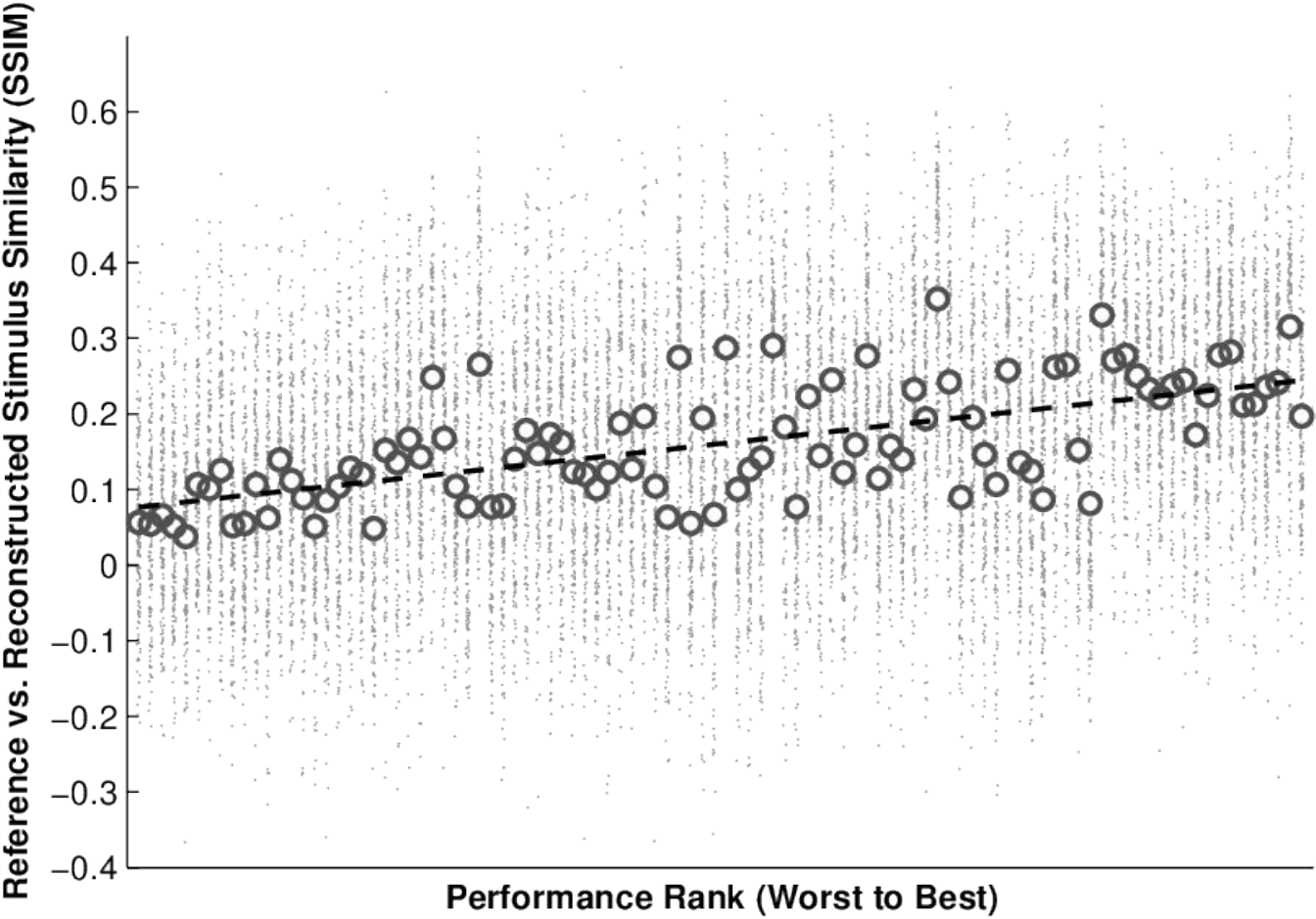}
\caption{ 
{\bf Encoding specificities vs.~performances of deep networks.} Each distribution from left to right shows the SSIM scores of the reconstructed stimuli of population representations of a deep network against the 16 reference stimuli. Performance ranks, instead of performance values, are used for visualization purposes. Means of distributions are plotted as gray circles and the linear regression as dashed black line. Significance of slope of means under permutation test has $p < 0.001$.}
\label{fig:SFenc}
\end{figure}

\end{document}